\documentclass{article}
\pdfoutput=1

     \PassOptionsToPackage{numbers, compress}{natbib}

    \usepackage[final]{neurips_2022}

\usepackage[utf8]{inputenc} 
\usepackage[T1]{fontenc}    
\usepackage{hyperref}       
\usepackage{url}            
\usepackage{booktabs}       
\usepackage{amsfonts}       
\usepackage{nicefrac}       
\usepackage{microtype}      
\usepackage{xcolor}         

\usepackage{graphicx}
 \graphicspath{{figures/}} 
\usepackage{subfigure}
\usepackage[figuresright]{rotating}
\usepackage{multicol}
\usepackage{multirow}

\usepackage{amsmath}
\usepackage{amssymb}
\usepackage{mathtools}
\usepackage{amsthm}

\usepackage{wrapfig}        
\usepackage{algorithm}
\usepackage{algorithmic}

\theoremstyle{plain}
\newtheorem{theorem}{Theorem}[section]

\theoremstyle{definition}
\newtheorem{definition}[theorem]{Definition}

\theoremstyle{remark}

\hypersetup{
	colorlinks=true
}

\title{Double Check Your State Before Trusting It: Confidence-Aware Bidirectional Offline Model-Based Imagination}

\author{%
  Jiafei Lyu$^{1}$, Xiu Li$^{1}$\thanks{Corresponding Authors}, Zongqing Lu$^{2 *}$ \\
  $^{1}$Tsinghua Shenzhen International Graduate School, Tsinghua University\\
  $^{2}$School of Computer Science, Peking University \\
  \texttt{lvjf20@mails.tsinghua.edu.cn,} \\
  \texttt{li.xiu@sz.tsinghua.edu.cn, zongqing.lu@pku.edu.cn}
}

\begin{document}

\maketitle

\begin{abstract}
  The learned policy of model-free offline reinforcement learning (RL) methods is often constrained to stay within the support of datasets to avoid possible dangerous out-of-distribution actions or states, making it challenging to handle out-of-support region. Model-based RL methods offer a richer dataset and benefit generalization by generating imaginary trajectories with either trained forward or reverse dynamics model. However, the imagined transitions may be inaccurate, thus downgrading the performance of the underlying offline RL method. In this paper, we propose to augment the offline dataset by using trained bidirectional dynamics models and rollout policies with \emph{double check}. We introduce conservatism by trusting samples that the forward model and backward model agree on. Our method, \textit{confidence-aware bidirectional offline model-based imagination}, generates reliable samples and can be combined with any model-free offline RL method. Experimental results on the D4RL benchmarks demonstrate that our method significantly boosts the performance of existing model-free offline RL algorithms and achieves competitive or better scores against baseline methods.
\end{abstract}

\section{Introduction}
Offline reinforcement learning (offline RL), also known as batch RL \cite{Lange2012BatchRL}, aims at learning from a static dataset that was previously gathered by an unknown behavioral policy. Offline RL is deemed to be promising \cite{Fujimoto2019OffPolicyDR, Fu2020D4RLDF} as online learning requires the agent to continuously interact with the environment, which however may be costly, time-consuming, or even dangerous. The progress in offline RL will undoubtedly scale RL methods to being widely applied in real-world applications, considering the impressive success in computer vision or natural language processing by adopting large-scale offline datasets \cite{Deng2009ImageNetAL, Chelba2014OneBW}.

Prior off-policy online RL methods \cite{Fujimoto2018AddressingFA, Haarnoja2018SoftAO, Munos2016SafeAE} are known to fail on fixed offline datasets, even on expert demonstrations \cite{Fu2020D4RLDF}, due to extrapolation errors \cite{Fujimoto2019OffPolicyDR}. In the offline setting, the agent can overgeneralize from the static dataset, resulting in arbitrarily wrong estimates upon out-of-distribution (OOD) state-action pairs and dangerous action execution. To address this issue, recent model-free offline RL algorithms compel the learned policy to stay close to the behavioral policy \cite{Fujimoto2019OffPolicyDR, Kumar2019StabilizingOQ, Wu2019BehaviorRO}, or incorporate some penalties into the critic \cite{Nachum2019AlgaeDICEPG, Kumar2020ConservativeQF, Kostrikov2021OfflineRL}. However, such approaches often suffer from loss of generalization capability \cite{Wu2021UncertaintyWA, Wang2021OfflineRL}, since they purposely avoid OOD states or actions.

Model-based offline RL methods, instead, enrich the logged dataset by generating synthetic samples with the trained forward or reverse (backward) dynamics model \cite{Kidambi2020MOReLM, Yu2020MOPOMO, Wang2021OfflineRL, Yu2021COMBOCO}. These methods benefit from better generalization thanks to richer transition samples. Intuitively, the performance of the agent is largely confined by the quality of the model-generated data, i.e., learning on bad states or actions will negatively affect the policy via backpropagation. Unfortunately, there is no guarantee that reliable transitions can be generated by the trained forward or backward dynamics model \cite{Asadi2018TowardsAS}.

\begin{wrapfigure}{r}{6.5cm}
    \centering
    \includegraphics[scale=0.47]{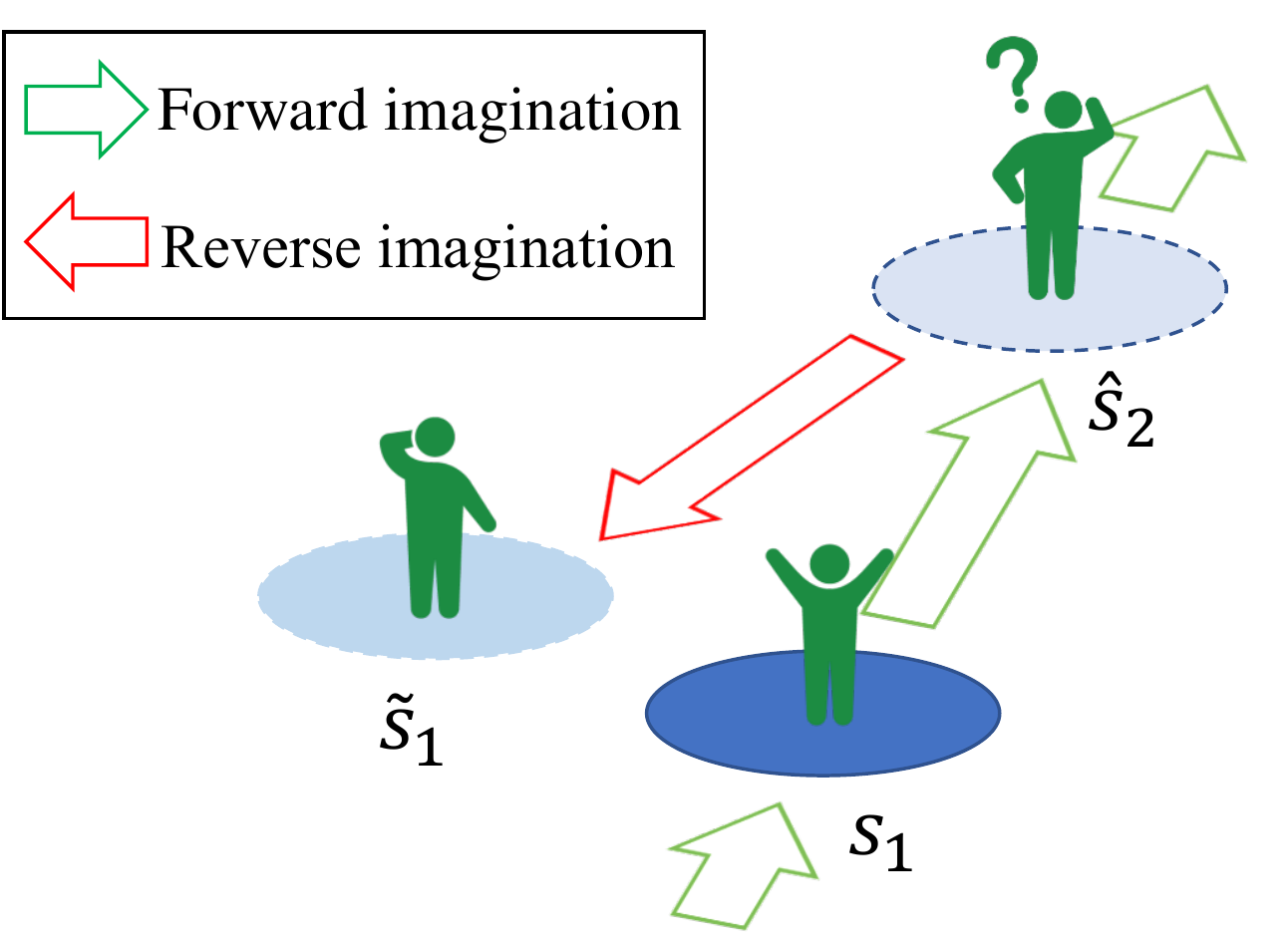}
    \vspace{-0.2cm}
    \caption{Illustration of the basic idea of confidence-aware bidirectional offline model-based imagination. To ensure that the forward imagination $\hat{s}_2$ from $s_1$ is valid and reasonable, one needs to `look back' to check whether the imagined previous state $\tilde{s}_1$ based on $\hat{s}_2$ is similar to $s_1$. We trust $\hat{s}_2$ if the deviation  between $s_1$ and $\tilde{s}_1$ is small.}
    \label{fig:doublecheck}
    \vspace{-0.2cm}
\end{wrapfigure}

In this paper, we aim to generate reliable transitions for offline RL via a \emph{double check mechanism}. The intuition behind this lies in the fact that humans often do double check when they are uncertain and need to be cautious. Besides the forward model, we train the backward model to generate simulated rollouts backward and use one to check whether the synthetic samples the other generated are credible. To be specific, we train bidirectional dynamics models along with bidirectional rollout policies. Instead of injecting pessimism into value estimation, we introduce \emph{conservatism into transition}, i.e., only samples that the forward model and reverse model agree on are trusted.

We use Figure \ref{fig:doublecheck} to further illustrate our insight, where we take forward transition generation as an example, of which the process is identical to the reverse setting. Starting from $s_1$, the forward model predicts next state $\hat{s}_2$. However, it is hard for the agent to decide whether $\hat{s}_2$ is trustworthy. One natural solution, which follows human's way of reasoning \cite{Holyoak1999BidirectionalRI}, is backtracking where it comes from, i.e., looking backward to trace previous state $\tilde{s}_1$, and check whether the imagined state $\tilde{s}_1$ based on $\hat{s}_2$ is different from the true state $s_1$. We are confident to $\hat{s}_2$ if $\tilde{s}_1$ is similar to $s_1$ and vice versa.

To this end, we propose \underline{C}onfidence-\underline{A}ware \underline{B}idirectional Offline Model-Based \underline{I}magination (CABI), which is a simple yet effective data augmentation method. CABI generally guarantees the reliability of the generated samples via the double check mechanism, and can be combined with any off-the-shelf model-free offline RL methods, e.g., BCQ \cite{Fujimoto2019OffPolicyDR} and TD3\_BC \cite{Fujimoto2021AMA}, to enjoy better generalization in a conservative manner. Extensive experimental results on the D4RL benchmarks \cite{Fu2020D4RLDF} show that CABI significantly boosts the performance of the base model-free offline RL methods, and achieves competitive or even better scores against recent model-free and model-based offline RL methods.

\section{Related Work}
In this paper, we consider offline reinforcement learning \cite{Lange2012BatchRL, Levine2020OfflineRL}, which defines the task of learning from a static dataset that was collected by an unknown behavior policy. Applications of offline RL include robotics \cite{Mandlekar2020IRISIR, Singh2020COGCN, Rafailov2021OfflineRL}, healthcare \cite{Gottesman2018EvaluatingRL, Wang2018SupervisedRL}, recommendation system \cite{Strehl2010LearningFL, Swaminathan2015BatchLF}, etc.

\noindent \textbf{Model-free offline RL.} Since it is risky to execute out-of-support actions, existing offline RL algorithms are often designed to constrain the policy search within the support of the static offline dataset. They realize it via importance sampling \cite{Precup2001OffPolicyTD, Sutton2016AnEA, Liu2019OffPolicyPG, Nachum2019DualDICEBE, Gelada2019OffPolicyDR}, explicit or implicit policy constraints \cite{Fujimoto2019OffPolicyDR, Kumar2019StabilizingOQ, Wu2019BehaviorRO, Laroche2019SafePI, Liu2020ProvablyGB, Zhou2020PLASLA}, learning conservative critics \cite{Kumar2020ConservativeQF, Ma2021ConservativeOD, Kumar2021DR3VD, Ma2021OfflineRL, Kostrikov2021OfflineRL, lyu2022mildly}, and quantifying estimation uncertainty \cite{Wu2021UncertaintyWA, Zanette2021ProvableBO, Deng2021SCORESC}. Recently, sequential modeling is also explored in the offline RL setting \cite{Chen2021DecisionTR, Janner2021ReinforcementLA, Meng2021OfflinePM}. Despite these advances, model-free offline RL methods suffer from loss of generalization beyond the dataset \cite{Wang2021OfflineRL}, and CABI is proposed to mitigate it.

\noindent \textbf{Model-based offline RL.} Model-based offline RL algorithms benefit from better generalization as the static dataset is extended by the synthetic samples generated from the trained forward \cite{Ross2012AgnosticSI, Finn2017DeepVF, Argenson2021ModelBasedOP} or reverse dynamics model \cite{Wang2021OfflineRL}. These methods heavily rely on uncertainty quantification \cite{Ovadia2019CanYT, Yu2020MOPOMO, Kidambi2020MOReLM, Diehl2021UMBRELLAUM}, compelling the policy towards the behavior policy \cite{Swazinna2021OvercomingMB, matsushima2021deploymentefficient}, representation learning \cite{lee2021representation, Rafailov2021OfflineRL}, and penalizing Q-values \cite{Yu2021COMBOCO}. However, it is hard to judge whether the transitions generated by the trained dynamics model are reliable, and poor imagined samples will negatively affect the performance of the agent. Recently, \cite{Yu2020MOPOMO, Lu2022RevisitingDC} explore how inaccurate rollouts that are well penalized can still be useful for model-based training. However, they work only for model-based offline RL methods. There are also some studies that focus on trajectory pruning \cite{Kidambi2020MOReLM, Zhan2021ModelBasedOP}, while they involve uncertainty measurement. CABI, instead, ensures reliable imaginations by conducting double check with the forward and backward models, which fully exploits the advantages of bidirectional modeling.

\noindent \textbf{Model-based online RL.} Model-based online RL methods achieve superior sample efficiency \cite{Sutton1990IntegratedAF, Kaelbling1996ReinforcementLA, Buckman2018SampleEfficientRL, Janner2019WhenTT} by learning a dynamics model of the environment and planning with the model \cite{Sutton2005ReinforcementLA, Walsh2010IntegratingSP, Wang2020Exploring}. Learning a backward dynamics model that produces traces towards the aimed state is also widely explored \cite{Edwards2018ForwardBackwardRL, pmlrv119lee20g, goyal2018recall, Lai2020BidirectionalMP}. Among them, most similar to our work is \cite{Lai2020BidirectionalMP}, which leverages bidirectional model rollouts for reduced compounding error in the online setting. However, the main differences are: (1) CABI is proposed to augment the fixed dataset instead of performing policy optimization in a model-based way; (2) CABI interpolates a double check mechanism for reliable imaginations; (3) Model predictive control (MPC) \cite{Camacho2013model} is not involved in CABI.

\begin{figure}
    \centering
    \includegraphics[scale=0.6]{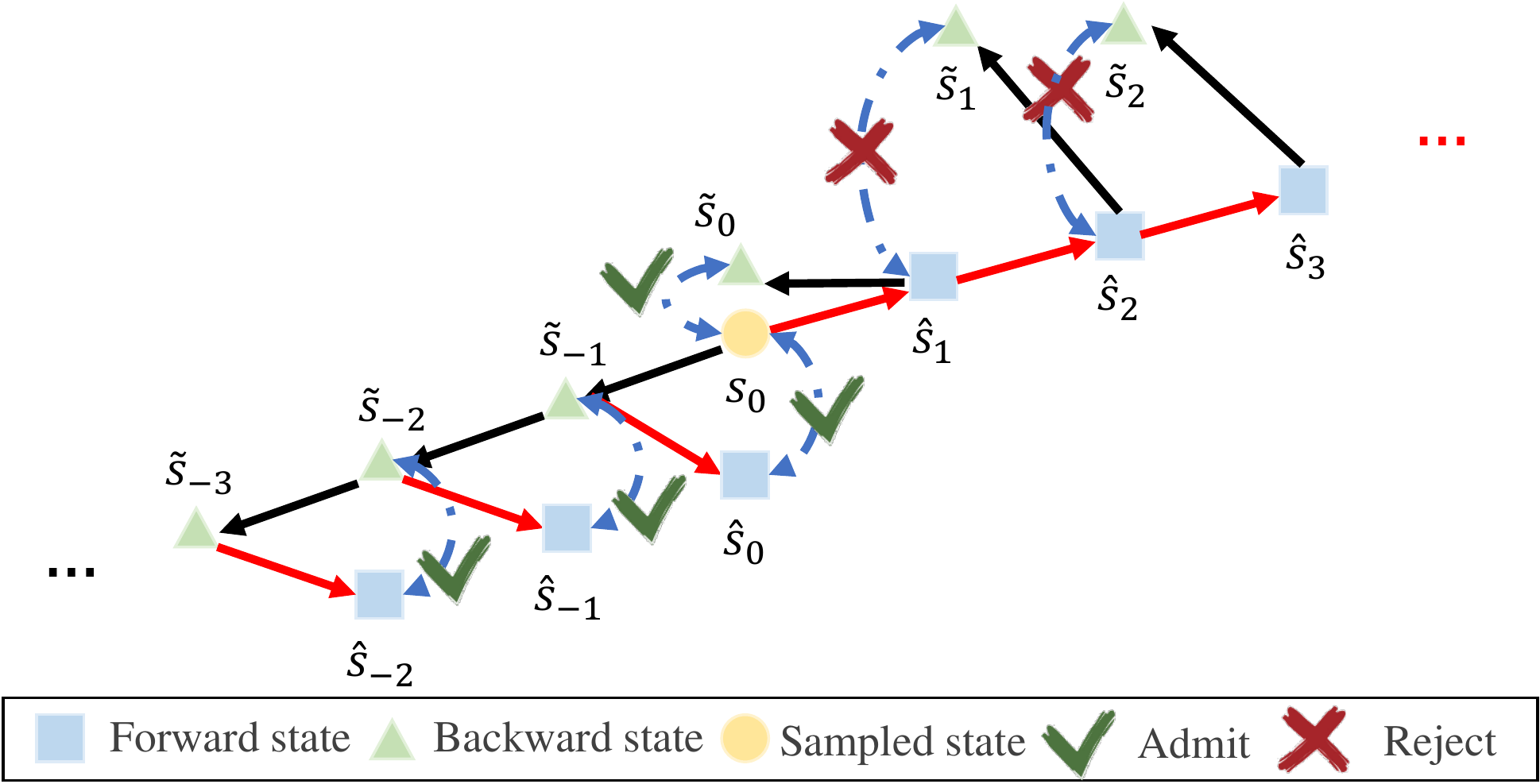}
    \caption{Illustration of the double check mechanism in bidirectional modeling. We use the trained opposite dynamics model to check whether the synthetic transitions generated by the current dynamics model are reliable. We are confident about the transitions that can be traced back to the starting state with small errors (green check) and reject those with large disagreements (red cross).}
    \label{fig:mechanism}
\end{figure}

\section{Preliminaries}
We study RL under Markov Decision Process (MDP) specified by a tuple $\langle \mathcal{S},\mathcal{A},\rho_0,p,r,\gamma \rangle$, where $\mathcal{S}$ is the state space, $\mathcal{A}$ is the action space, $\rho_0$ denotes initial state distribution, $p(s^\prime|s,a)$ is the stochastic transition dynamics, $r(s,a):\mathcal{S}\times \mathcal{A}\mapsto \mathbb{R}$ is the reward function, and $\gamma\in[0,1)$ is the discount factor. 
The policy $\pi(a|s): \mathcal{S}\times\mathcal{A}\mapsto \mathbb{R}_+$ is a mapping from states to a probability distribution over actions. The goal of RL is to obtain a policy $\pi$ such that the expected discounted cumulative rewards can be maximized, $\max_{\pi} J_{\rho_0}(\pi):= \mathbb{E}_{s\sim \rho_0,a_t\sim\pi(\cdot|s_t), s_{t+1}\sim p(\cdot|s_t,a_t)}\left[ \sum_{t=0}^\infty \gamma^t r(s_t, a_t) \right]$.
In online RL, the agent learns from the experience collected from the interactions with the environment. However, in the offline setting, both interaction and exploration are infeasible, and the agent can only get access to the logged static dataset $\mathcal{D}_{\mathrm{env}}=\{(s,a,r,s^\prime)\}$, which was gathered in advance by the unknown behavior policy. Since the fixed dataset $\mathcal{D}_{\mathrm{env}}$ is typically a subset of full space $\mathcal{S}\times\mathcal{A}$, the generalization beyond the raw dataset becomes challenging. Model-based RL mitigates this issue by learning a dynamics model $\hat{p}(s^\prime|s,a)$ and reward function $\hat{r}(s,a)$, and generating synthetic transitions to augment the dataset. However, there is no guarantee that the generated samples are reliable (see Section \ref{sec:checkyourstate}), and we focus on addressing this issue in this paper.

\section{Confidence-Aware Bidirectional Offline Model-Based Imagination}

In this section, we first use a toy example to illustrate the necessity of training bidirectional models with the double check mechanism. Then, we give the detailed framework of our method, \underline{C}onfidence-\underline{A}ware \underline{B}idirectional Offline Model-Based \underline{I}magination (CABI).

\subsection{You Need to Double Check Your State}
\label{sec:checkyourstate}

Many model-free offline RL algorithms suffer from poor generalization as they are trained on a fixed dataset with limited samples. Model-based methods extend the logged dataset by generating synthetic transitions from the trained dynamics model. Despite such an advantage, they lack a mechanism for checking the \emph{reliability} of the generated samples. If the model is inaccurate, poor transition samples that lie in the out-of-support region of the dataset can be generated, which may downgrade the performance of offline RL algorithms.

%
\begin{wrapfigure}{r}{4cm}
    \centering
    \setlength{\belowcaptionskip}{-0.1cm}
    \includegraphics[scale=0.35]{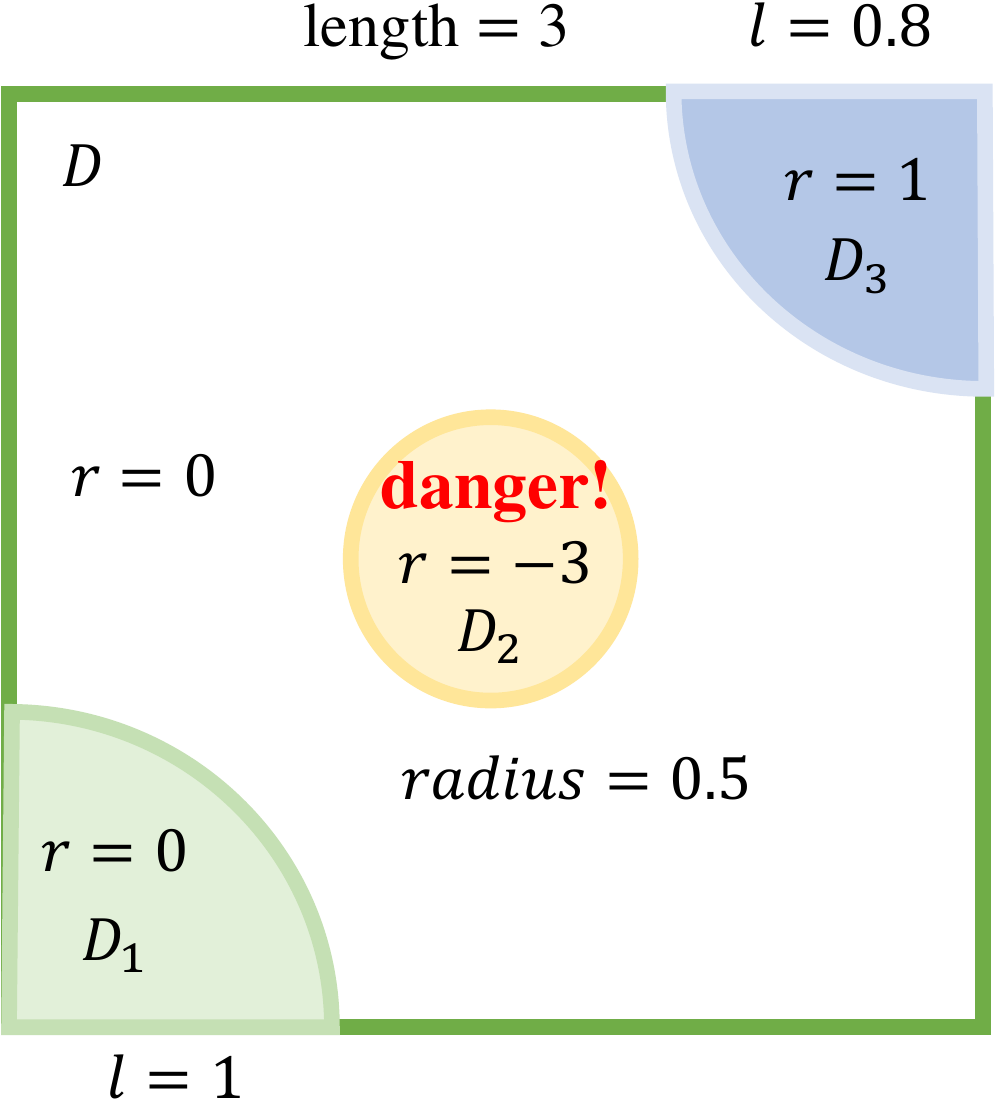}
    \caption{RiskWorld.}
    \label{fig:riskworld}
\end{wrapfigure}
Human beings tend to conduct \emph{double check} when they are uncertain about the outcome, e.g., clinical medicine research \cite{Subramanyam2016InfusionME}, autonomous driving \cite{Van2012ANA}, etc. Inspired by this nature, we propose to train bidirectional dynamics models and admit the samples where the forward model and backward model have few disagreements instead of roughly trusting all generated samples. In this way, we introduce conservatism into the transition itself instead of the critic or the actor networks. We give the illustration of double check mechanism in Figure \ref{fig:mechanism}. 

We argue that either forward dynamics model or backward dynamics model is unreliable, and bidirectional modeling in conjunction with the double check mechanism is critical for trustworthy sample generation. We verify this by designing a toy task, 2-dimensional environment with continuous state space and action space, namely RiskWorld, as shown in Figure \ref{fig:riskworld}. The central point of the square region in RiskWorld is $(0,0)$, and the state space gives $D \coloneqq [-1.5,1.5]\times[-1.5,1.5]$. Each episode, the agent randomly starts at the region $D_1 \coloneqq\{ (x,y) | (x+1.5)^2+(y+1.5)^2 \le 1, x <0, y<0 \}$ and takes actions $a\in[-0.5,0.5]$. There is a danger zone $D_2 \coloneqq \{ (x,y)| x^2+y^2 \le 0.5^2 \}$, and the done flag would turn into true if the agent steps into this region, along with a reward of $-3$. The agent will receive a reward of $+1$ if it lies in $D_3 \coloneqq \{ (x,y)| (x-1.5)^2+(y-1.5)^2 \le 0.8^2, x<1.5, y<1.5 \}$, and $0$ if it locates at $D\backslash (D_2\cup D_3)$.

\begin{figure*}[!h]
\setlength{\belowcaptionskip}{-0.1cm}
    \centering
    \subfigure[Random policy]{
    \label{fig:random}
    \includegraphics[scale=0.185]{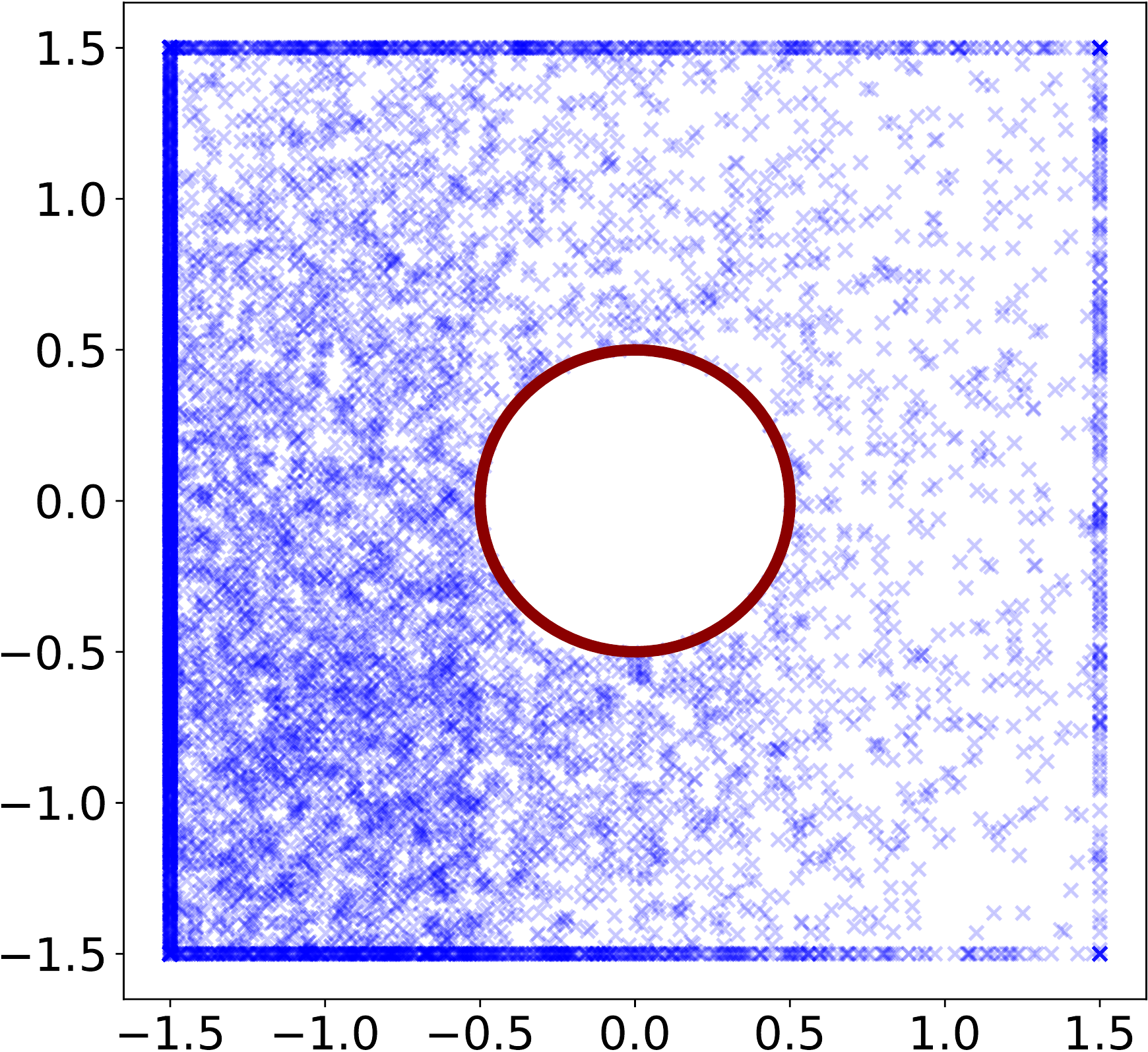}
    }\hspace{-2mm}
    \subfigure[Forward]{
    \label{fig:forward}
    \includegraphics[scale=0.185]{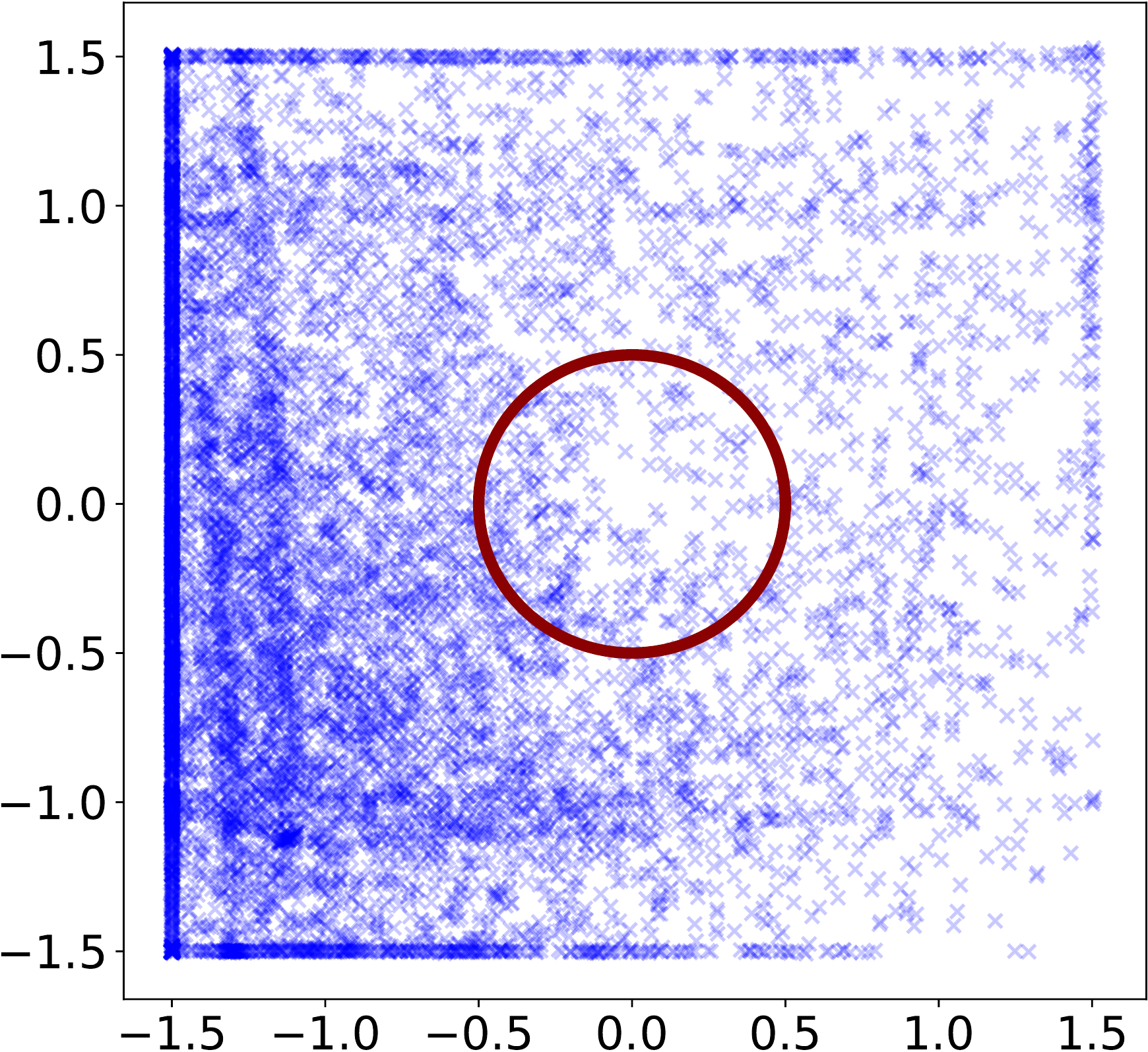}
    }\hspace{-2mm}
    \subfigure[Backward]{
    \label{fig:reverse}
    \includegraphics[scale=0.185]{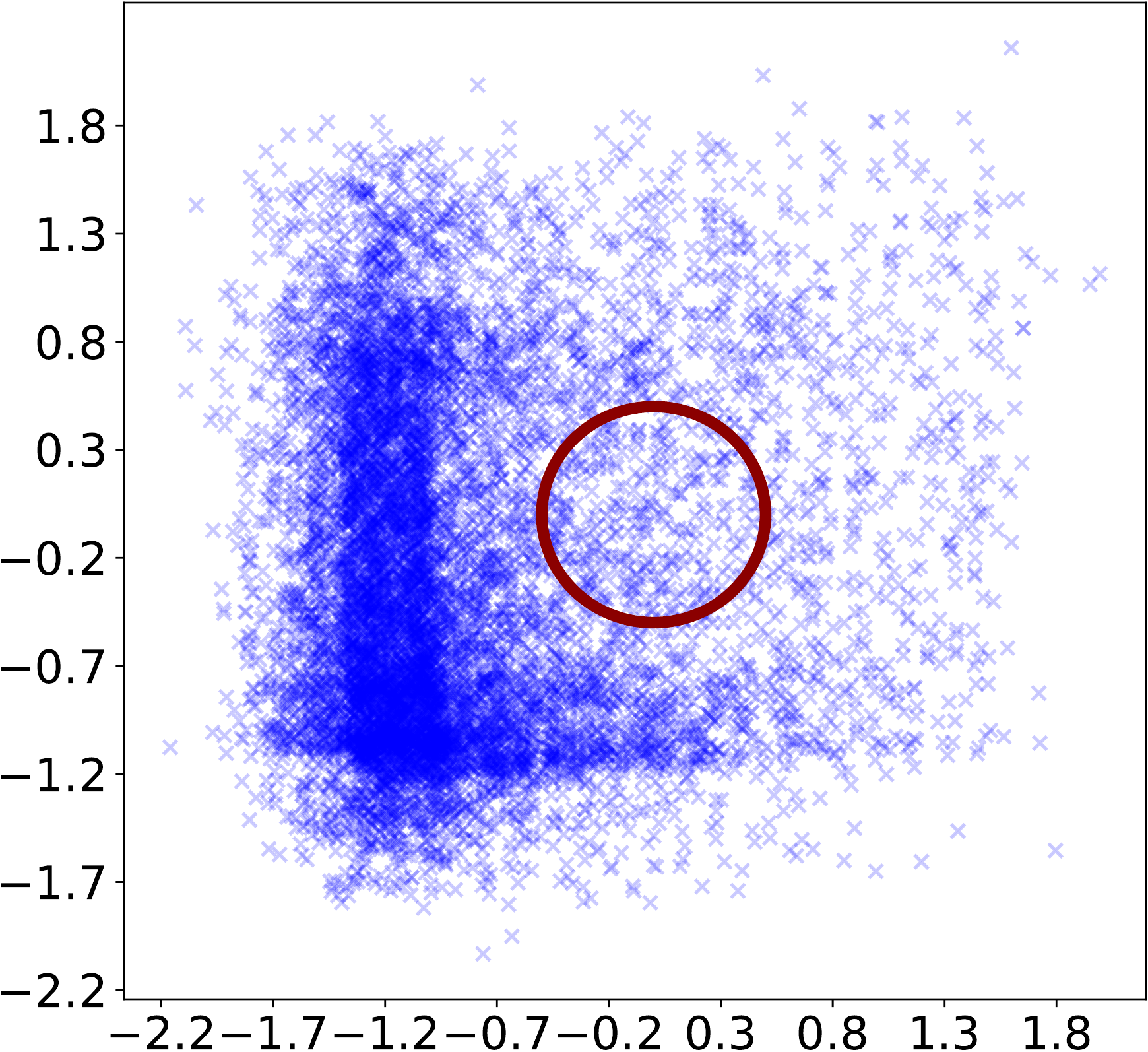}
    }\hspace{-2mm}
    \subfigure[Bidirectional]{
    \label{fig:bidirectional}
    \includegraphics[scale=0.185]{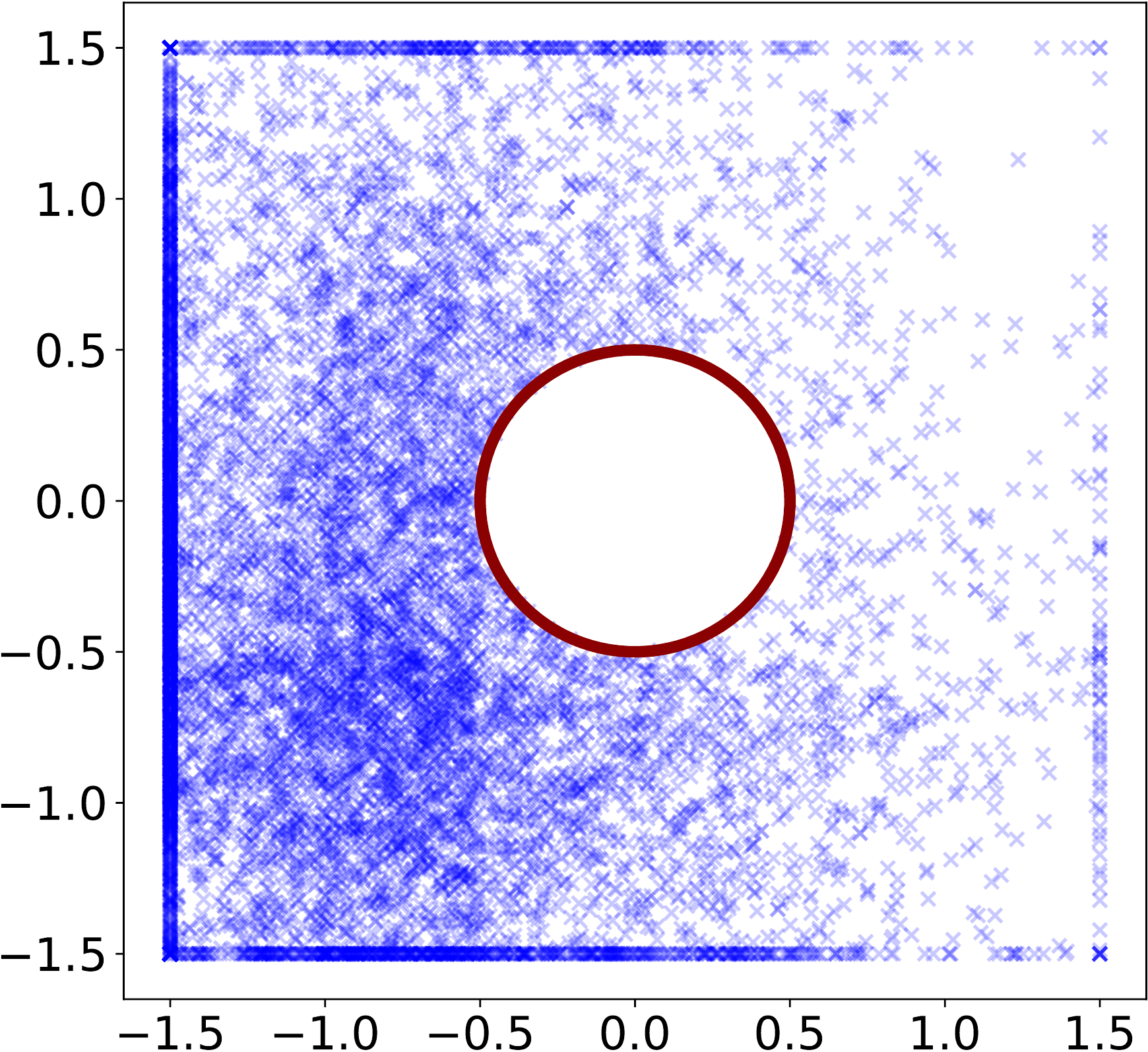}
    }
    \caption{Visualizations of the offline dataset collected by a random policy (a), and synthetic transitions generated by a forward model (b), a backward model (c), a bidirectional model with the double check mechanism (d), based on the offline dataset in (a).}
    \label{fig:toyexample}
\end{figure*}

We run a random policy in RiskWorld for $10^4$ timesteps to collect an offline dataset. Figure \ref{fig:random} shows the state distribution (blue cross) of the dataset, where there are no transitions in $D_2$ (red circle area) as the episode terminates if the agent steps into $D_2$. To compare different ways of imagination, we train a forward model, a backward model, and a bidirectional model with the double check mechanism on this dataset. The training epoch is set to be 100, and the rollout horizon is set to be 3 for all of them. The detailed experimental setup is available in Appendix \ref{sec:setupriskworld}.
\vspace{-2mm}

We use the trained forward model, reverse model, and bidirectional model to generate $10^4$ transition samples, and plot the state distributions of their generated samples respectively. As shown in Figure \ref{fig:forward}, the forward model generates many samples in the danger zone $D_2$ (red circle area). Figure \ref{fig:reverse} reveals that the reverse model generates a lot of illegal samples that lie out of the state space $D$, and also many transitions that lie in the dangerous area $D_2$. These all show evidence that both the forward and backward dynamics model fail to output reliable transitions. However, we observe in Figure \ref{fig:bidirectional} that bidirectional modeling with the double check mechanism successfully produces reliable and conservative synthetic samples, i.e., out-of-support or dangerous samples are not included, because the forward model and backward model have large disagreements at those states. We hence argue that bidirectional modeling with double check is necessary for reliable data generation in offline RL.

\subsection{Bidirectional Models Learning in CABI}
\label{sec:modellearning}

\noindent\textbf{Bidirectional dynamics models training.} Our bidirectional modeling models transition probability and reward function simultaneously, i.e., the forward model $\hat{p}_\psi(s^\prime, r|s,a)$ and the reverse model $\hat{p}_\phi(s,r|s^\prime,a)$ parameterized by $\psi$ and $\phi$ respectively. The forward model $\hat{p}_\psi(s^\prime, r|s,a)$ represents the probability of the next state and corresponding reward given the current state and action, and the backward model $\hat{p}_\phi(s,r|s^\prime,a)$ outputs the probability of the current state and reward using the next state and action as input. We assume that the predicted reward function $\hat{r}(s,a)$ only depends on the current state $s$ and action $a$, then the unified model can be decomposed as $\hat{p}_\psi(s^\prime, r|s,a) = \hat{p}(s^\prime|s,a)\hat{p}(r|s,a)$ and $\hat{p}_\phi(s,r|s^\prime,a) = \hat{p}(s|s^\prime,a)\hat{p}(r|s,a)$.
We denote the loss functions for the forward model and backward model as $\mathcal{L}_\psi^{\mathrm{fwd}}$ and $\mathcal{L}_\phi^{\mathrm{bwd}}$ respectively, and optimize them by maximizing the log-likelihood via \eqref{eq:forwardmodeltraining} and \eqref{eq:reversemodeltraining}, where $\mathcal{D}_{\mathrm{env}}$ is the raw static dataset. 
\begin{gather}
    \mathcal{L}_\psi^{\mathrm{fwd}} = \mathbb{E}_{(s,a,r,s^\prime)\sim\mathcal{D}_{\mathrm{env}}} \left[ -\log \hat{p}_\psi(s^\prime,r|s,a) \right], \label{eq:forwardmodeltraining} \\
    \mathcal{L}_\phi^{\mathrm{bwd}} = \mathbb{E}_{(s,a,r,s^\prime)\sim\mathcal{D}_{\mathrm{env}}} \left[ -\log \hat{p}_\phi(s,r|s^\prime,a) \right]. \label{eq:reversemodeltraining}
\end{gather}

\begin{algorithm}[tb]
\caption{CABI (Model Training)}
\label{alg:cabimodeltraining}
{\bf Input:} Dataset $\mathcal{D}_{\mathrm{env}}$, iterations $N_1, N_2$, learning rate $\alpha_\psi$, $\alpha_\phi$, $\alpha_\theta$, $\alpha_\omega$
\begin{algorithmic}[1]
\STATE Randomly initialize forward model parameters $\psi$ and reverse model parameters $\phi$ 
\FOR{$i$ = 1 to $N_1$}
\STATE Compute $\mathcal{L}_\psi^{\mathrm{fwd}}$ and $\mathcal{L}_\phi^{\mathrm{bwd}}$ via \eqref{eq:forwardmodeltraining} and \eqref{eq:reversemodeltraining}
\STATE Update model parameters: $\psi \leftarrow \psi - \alpha_\psi \nabla_\psi \mathcal{L}_\psi^{\mathrm{fwd}}$, $\phi \leftarrow \phi - \alpha_\phi \nabla_\phi \mathcal{L}_\phi^{\mathrm{bwd}}$
\ENDFOR
\STATE Randomly initialize forward and backward rollout policy parameters $\theta$, $\omega$
\FOR{$i$ = 1 to $N_2$}
\STATE Compute $\mathcal{L}_{\mathrm{fvae}}$ and $\mathcal{L}_{\mathrm{bvae}}$ via \eqref{eq:forwardvae} and \eqref{eq:reversevae}
\STATE Update forward and reverse rollout policy: $\theta \leftarrow \theta - \alpha_\theta \nabla_\theta \mathcal{L}_{\mathrm{fvae}}$, $\omega \leftarrow \omega - \alpha_\omega \nabla_\omega \mathcal{L}_{\mathrm{bvae}}$
\ENDFOR
\end{algorithmic}
\end{algorithm}

Following prior work \cite{Yu2020MOPOMO, Kidambi2020MOReLM}, we train an ensemble of bootstrapped probabilistic dynamics models, which has been widely demonstrated to be effective in model-based RL \cite{Chua2018DeepRL,Janner2019WhenTT}. Each model in the ensemble is parameterized by a multi-layer neural network, which outputs a Gaussian distribution $\mathcal{N}(\mu,\Sigma)$. Detailed hyperparameter setup for dynamics models training is deferred to Appendix \ref{sec:implementationdetails}.

\noindent\textbf{Bidirectional rollout policies training.} We additionally train bidirectional generative models, which serve as rollout policies, and are used to generate actions to augment the static dataset. We model the rollout policy with a conditional variational autoencoder (CVAE) \cite{Kingma2014AutoEncodingVB,Sohn2015LearningSO, Fujimoto2019OffPolicyDR}, which offers diverse actions while staying within the span of the dataset. CVAE is made up of an encoder ${E}$ that outputs the latent variable $z$ under the Gaussian distribution, and a decoder ${D}$ that maps $z$ to the desired space. We denote the forward rollout policy as ${G}_\theta^{\mathrm{fwd}}(s)$ parameterized by $\theta=\{ \xi_1, \nu_1\}$ where $\xi_1$ is the parameter of the encoder ${E}_{\xi_1}^{\mathrm{fwd}}(s,a)$ and $\nu_1$ is the parameter of the decoder ${D}_{\nu_1}^{\mathrm{fwd}}(s,z)$. The forward rollout policy is then trained by maximizing its variational lower bound, which is equivalent to minimizing the following loss: \vspace{-0.2cm}
\begin{equation}
\label{eq:forwardvae}
\begin{aligned}
    \displaystyle \mathcal{L}_{\mathrm{fvae}} (\theta) = \mathop{\mathbb{E}}\limits_{\substack{ (s,a,r,s^\prime)\sim\mathcal{D}_{\mathrm{env}} , z\sim{E}_{\xi_1}^{\mathrm{fwd}}(s,a)}} [ \left( a - {D}_{\nu_1}^{\mathrm{fwd}}(s,z) \right)^2 + \left . D_{\mathrm{KL}}\left( {E}_{\xi_1}^{\mathrm{fwd}}(s,a) \| \mathcal{N}(0,\bf{I}) \right) \right],
\end{aligned}
\end{equation}
where $D_{\mathrm{KL}}(\cdot\| \cdot)$ denotes the KL-divergence, and $\bf{I}$ is an identity matrix. The first term of RHS of \eqref{eq:forwardvae} represents the reconstruction loss where we want the decoded action to approximate the real action. Then for action generation, we first sample latent vector $z$ from the multivariate Gaussian distribution $\mathcal{N}(0,\bf{I})$, and then pass it with the current state $s$ into the decoder ${D}_{\nu_1}^{\mathrm{fwd}}(s,z)$ to output the action.

Similarly, the backward rollout policy ${G}_{\omega}^{\mathrm{bwd}}(s^\prime)$ parameterized by $\omega$ contains an encoder ${E}_{\xi_2}^{\mathrm{bwd}}(s^\prime,a)$ and a decoder ${D}_{\nu_2}^{\mathrm{bwd}}(s^\prime,z)$, $\omega=\{ \xi_2,\nu_2 \}$. The loss function of the backward rollout policy gives:
\begin{equation}
\label{eq:reversevae}
\begin{aligned}
    \displaystyle \mathcal{L}_{\mathrm{bvae}} (\omega) = \mathop{\mathbb{E}}\limits_{\substack{(s,a,r,s^\prime)\sim\mathcal{D}_{\mathrm{env}} , z\sim{E}_{\xi_2}^{\mathrm{bwd}}(s^\prime,a)}} [ \left( a - {D}_{\nu_2}^{\mathrm{bwd}}(s^\prime,z) \right)^2  + \left . D_{\mathrm{KL}}\left( {E}_{\xi_2}^{\mathrm{bwd}}(s^\prime,a) \| \mathcal{N}(0,\bf{I}) \right) \right].
\end{aligned}
\end{equation}
We then draw $z$ from the Gaussian distribution $\mathcal{N}(0,\bf{I})$, and draw the action from the action decoder ${D}_{\nu_2}^{\mathrm{bwd}}(s^\prime,z)$ with the next state $s^\prime$ and latent variable $z$ as input. 

We present the detailed procedure for the model training part of CABI in Algorithm \ref{alg:cabimodeltraining}.

\subsection{Conservative Data Augmentation with CABI}
\label{sec:dataaugmentation}
After the bidirectional dynamics models and bidirectional rollout policies are well trained, we utilize them to generate imaginary samples. Each time, we sample a state $s_t$ from the raw dataset $\mathcal{D}_{\mathrm{env}}$ to produce imagined forward trajectory $\hat{\tau}^{\mathrm{fwd}} = \{s_{t+j},a_{t+j},r_{t+j},s_{t+1+j}\}_{j=0}^{H-1}$ with the forward dynamics model $\hat{p}_\psi$ and forward rollout policy ${G}_{\theta}^{\mathrm{fwd}}$, and sample the next state $s_{t+1}$ from $\mathcal{D}_{\mathrm{env}}$ to generate imagined reverse trajectory $\hat{\tau}^{\mathrm{bwd}} = \{ s_{t-j}, a_{t-j}, r_{t-j}, s_{t+1-j} \}_{j=0}^{H-1}$ with the reverse dynamics model $\hat{p}_\phi$ and reverse rollout policy ${G}_{\omega}^{\mathrm{bwd}}$. For each step in the rollout horizon $H$, we do double check and reject those badly imagined synthetic transitions. 

To be specific, when performing forward imagination from $s_t$ and generating synthetic next state $\hat{s}_{t+1}$, we trace back from $\hat{s}_{t+1}$ with the reverse model, and get the backward state $\tilde{s}_t$. We evaluate the deviation of $\tilde{s}_t$ from $s_t$, and trust $\hat{s}_{t+1}$ if the deviation is small. Similarly, starting from the state $s_{t+1}$, we backtrack its previous state $\tilde{s}_t$ with the backward dynamics model, and then look forward from $\tilde{s}_t$ to get $\hat{s}_{t+1}$ with the forward dynamics model. We trust $\tilde{s}_t$ if the deviation of $\hat{s}_{t+1}$ from $s_{t+1}$ is small.

We keep those trustworthy rollouts and gather them to get the model buffer $\mathcal{D}_{\mathrm{model}}$. We combine the synthetic model buffer $\mathcal{D}_{\mathrm{model}}$ with $\mathcal{D}_{\mathrm{env}}$ to obtain the final buffer $\mathcal{D}_{\mathrm{total}}$, i.e., $\mathcal{D}_{\mathrm{total}} = \mathcal{D}_{\mathrm{env}}\cup \mathcal{D}_{\mathrm{env}}$. We then can train \emph{any} model-free offline RL algorithms based on the composite dataset.

One na\"ive way for implementing double check mechanism is to set a threshold $\delta$, and admit the transition if $\| s_t-\tilde{s}_t \|_2 \le \delta$ for forward imagination, or $\| s_{t+1} - \hat{s}_{t+1} \|_2\le\delta$ for backward imagination. However, such a method lacks flexibility, and one may need to carefully tune $\delta$ per dataset based on the strong prior knowledge about the dataset, which impedes the application of double check mechanism. We resort to sorting the transitions in a mini-batch by the state deviation from small to large and keep the top $k$\% of them that have the smallest deviation. We keep 20\% transitions that have the smallest deviation for all of our experiments in Section \ref{sec:experiment} (empirical study on $k$ is available in Appendix \ref{sec:implementationdetails}).

Our method is confidence-aware and conservative as we only admit the transitions that the forward model and backward model agree on, thus excluding those poor transitions from the model buffer $\mathcal{D}_{\mathrm{model}}$. The full procedure for the data generation part of CABI is available in Algorithm \ref{alg:cabidatageneration}.

\begin{algorithm}[!t]
\caption{CABI (Data Generation)}
\label{alg:cabidatageneration}
{\bf Input:} Offline dataset $\mathcal{D}_{\mathrm{env}}$, horizon $H$, iteration $N$

\begin{algorithmic}[1] 
\STATE Initialize model replay buffer $\mathcal{D}_{\mathrm{model}}\leftarrow \emptyset$
\FOR{$i$ = 1 to $N$}
\STATE Sample state $s_t$ and next state $s_{t+1}$ from $\mathcal{D}_{\mathrm{env}}$
\FOR{$j$ = 0 to $H-1$}
\STATE Obtain forward rollout $\{ s_{t+j}, a_{t+j}, r_{t+j}, s_{t+1+j}\}$ from $s_{t+j}$ by drawing samples from the forward model $\hat{p}_{\psi}$ and forward rollout policy ${G}_\theta^{\mathrm{fwd}}$
\STATE Generate backward state $\tilde{s}_{t+j}$ from $s_{t+1+j}$, and evaluate the deviation of $\tilde{s}_{t+j}$ from $s_{t+j}$
\STATE Get backward rollout $\{ s_{t-j}, a_{t-j},r_{t-j}, s_{t+1-j}\}$ from $s_{t+1-j}$ by drawing samples from the backward model $\hat{p}_{\phi}$ and backward rollout policy ${G}_\omega^{\mathrm{bwd}}$
\STATE Generate forward state $\hat{s}_{t+1-j}$ from $s_{t-j}$. Evaluate the deviation of $\hat{s}_{t+1-j}$ from $s_{t+1-j}$
\STATE Add selected imaginations into $\mathcal{D}_{\mathrm{model}}$
\ENDFOR
\ENDFOR
\STATE Get composite dataset $\mathcal{D}_{\mathrm{total}} \leftarrow \mathcal{D}_{\mathrm{env}}\bigcup \mathcal{D}_{\mathrm{model}}$
\STATE Get the final policy $\pi_{\mathrm{out}}$ with \emph{any} model-free offline RL algorithm based on the dataset $\mathcal{D}_{\mathrm{total}}$
\end{algorithmic}
\end{algorithm}

\begin{table*}[!t]
  \caption{Normalized average score comparison of CABI+BCQ against different baselines on the Adroit "-v0" tasks, where score 0 represents the performance of a random policy and 100 corresponds to an expert policy performance. The results are averaged over the final 10 evaluations and 5 different random seeds. The highest mean scores are in \textbf{bold}.}
  \vspace{0.2cm}
  \renewcommand\arraystretch{1.05}
  \label{tab:adroitscorecompare}
  \centering
  \footnotesize
  \begin{tabular}{@{}llllllllll@{}}
    \toprule
    Task Name  & CABI+BCQ & BCQ & UWAC & BEAR & BC & AWR & CQL & MOPO & COMBO \\
    \midrule
    pen-cloned  & 54.7$\pm$2.0 & 44.0 & 33.1 & 26.5 & \textbf{56.9} & 28.0 & 39.2 & -2.1 & -2.4\\
    pen-human & \textbf{75.1}$\pm${1.5} & 68.9& 21.7 & -1.0 & 34.4 & 12.3 & 37.5 & 9.7  & 27.7\\
    pen-expert & \textbf{127.6}$\pm${2.0} & 114.9 & 111.9 & 105.9 & 85.1 & 111.0 & 107.0 & -0.6 & 11.5  \\
    door-cloned & \textbf{0.5}$\pm$0.2 & 0.0 & 0.0 & -0.1 & -0.1 & 0.0 & 0.4 & -0.1  & 0.0   \\
    door-human & 1.7$\pm$0.1 & 0.0 & 2.1 & -0.3 & 0.5 & 0.4 & \textbf{9.9} & -0.2  & -0.3 \\
    door-expert & \textbf{105.3}$\pm${0.5} & 99.0 & 104.1 & 103.4 & 34.9 & 102.9 & 101.5 & -0.2  & 4.9  \\
    relocate-cloned & -0.2$\pm$0.0 & -0.3 & -0.3 & -0.3 & \textbf{-0.1} & -0.2 & \textbf{-0.1} & -0.3  & \textbf{-0.1}   \\
    relocate-human & 0.1$\pm$0.1 & \textbf{0.5} & \textbf{0.5} & -0.3 & 0.0 & 0.0 & 0.2 & -0.3  & -0.3  \\
    relocate-expert & \textbf{105.9}$\pm${1.0} & 41.6 & 105.6 & 98.6 & 101.3 & 91.5 & 95.0 & -0.2  & 17.2  \\
    hammer-cloned & \textbf{4.3} $\pm${1.6} & 0.4 & 0.4 & 0.3 & 0.8 & 0.4 & 2.1 & 0.2 & 0.4  \\
    hammer-human & 3.1$\pm$2.2 & 0.5 & 1.1 & 0.3 & 1.5 & 1.2 & \textbf{4.4} & 0.2 & 0.2   \\
    hammer-expert & \textbf{128.9}$\pm${0.9} & 107.2 & 110.6 & 127.3 & 125.6 & 39.0 & 86.7 & 0.3  & 0.3  \\
    \midrule
    Total Score & \textbf{607.0} & 476.7 & 490.8 & 460.3 & 440.8 & 386.5 & 483.8 & 6.4 & 59.1 \\
    \bottomrule
  \end{tabular}
\end{table*}

\section{Experiments}
\label{sec:experiment}
In this section, we combine CABI with off-the-shelf model-free offline RL algorithms and conduct extensive experiments on the D4RL benchmarks \cite{Fu2020D4RLDF}. In Section \ref{sec:adroit}, we combine CABI with BCQ \cite{Fujimoto2019OffPolicyDR}, and evaluate it on the challenging Adroit dataset to show the effectiveness of conservative data augmentation with CABI. We present a detailed ablation study in Section \ref{sec:ablation}, where we aim to answer the following questions: (1) Is the double check mechanism a critical component for CABI? (2) How does CABI compare with the forward/reverse imagination? (3) How does CABI compare against other augmentation methods, e.g., random selection? Furthermore, we incorporate CABI with another recent model-free offline RL method, TD3\_BC \cite{Fujimoto2021AMA}, and evaluate it on the MuJoCo datasets, to show the generality and advantages of CABI. We additionally combine CABI with IQL \cite{kostrikov2022offline} and evaluate the performance of CABI+IQL on both Adroit tasks and MuJoCo tasks. Due to the space limit, the results are deferred to Appendix \ref{sec:iql}.

\subsection{Performance on Challenging Adroit Dataset}
\label{sec:adroit}

We demonstrate the benefits of CABI by combining it with BCQ and evaluating it on the challenging Adroit dataset \cite{Rajeswaran2018LearningCD}. Adroit dataset involves controlling a 24-DoF simulated robotic hand that aims at hammering a nail, opening a door, twirling a pen, or picking/moving a ball. It contains three types of datasets for each task (\emph{human}, \emph{cloned}, and \emph{expert}), yielding a total of 12 datasets. This domain is very challenging for prior methods to learn from because the dataset is made up of narrow human demonstrations on a sparse reward, high-dimensional robotic manipulation task.

We summarize the overall results in Table \ref{tab:adroitscorecompare}, where we compare CABI+BCQ against recent model-free offline RL methods, such as UWAC \cite{Wu2021UncertaintyWA}, CQL \cite{Kumar2020ConservativeQF}, BCQ \cite{Fujimoto2019OffPolicyDR}, and model-based offline RL methods, such as MOPO \cite{Yu2020MOPOMO}, and COMBO \cite{Yu2021COMBOCO}. We run MOReL and COMBO on these datasets with our reproduced code. Results of MOPO and UWAC on the Adroit domain are acquired by running their official codebases, and the results of the rest baselines are taken directly from \cite{Fu2020D4RLDF}. All methods are run over 5 different random seeds and normalized average scores are reported in Table \ref{tab:adroitscorecompare}. We only report the standard deviation for CABI+BCQ, and the full table is deferred to Appendix \ref{sec:fullcomptable}.

As shown, CABI significantly boosts the performance of vanilla BCQ on almost all datasets, achieving a total score of \textbf{607.0} vs. 476.7 of BCQ. CABI+BCQ also surpasses the baseline model-free and model-based offline RL methods on 7 out of 12 datasets and achieves the highest total score.

It is worth noting that model-based offline RL methods generally fail on the Adroit tasks, because (1) the dataset distribution is narrow and high-dimensional, making it challenging for the trained forward dynamics model to generate accurate and reliable transitions; (2) the actions in the synthetic transitions are generated by the actor during the training process, thus the error may accumulate if the actor is updated towards a wrong direction. CABI, instead, alleviates the underlying issues via adopting the CVAE for action generation and conducting double check on state prediction.

\subsection{Ablation Study}
\label{sec:ablation}

\noindent\textbf{Is the double check mechanism critical?} To answer this question, we exclude the double check mechanism in CABI and admit all generated synthetic samples from bidirectional models, which gives rise to Bidirectional Offline Model-based Imagination (BOMI). We evaluate CABI+BCQ and BOMI+BCQ on the Adroit tasks with identical parameter configuration over 5 different random seeds and show the average normalized scores in Table \ref{tab:wodoublecheckmechanism}. It can be seen that BOMI brings some performance improvement on most of the tasks via data augmentation with bidirectional models and rollout policies. However, the generated data may be unreliable (we observe a performance drop in \emph{pen-cloned, pen-human}), which impedes the benefits of bidirectional data augmentation. Such concern can be alleviated with the aid of the double check mechanism. As illustrated in Table \ref{tab:wodoublecheckmechanism}, CABI+BCQ outperforms BOMI+BCQ on most tasks and incurs a much better total score. 


\begin{table}[t]
  \caption{Normalized average score comparison on the Adroit tasks between CABI+BCQ, forward imagination+BCQ, backward imagination+BCQ, and BOMI+BCQ, where $\pm$ captures standard deviation. The results are averaged over the final 10 evaluations and 5 different random seeds. The highest mean scores are in \textbf{bold}.}
  \vspace{0.2cm}
  \renewcommand\arraystretch{1.05}
  \label{tab:wodoublecheckmechanism}
  \centering
  \small
  \begin{tabular}{llllll}
    \toprule
    Task name  & BCQ & +Forward & +Backward & +BOMI & +CABI \\
    \midrule
    pen-cloned & 44.0 & 41.2$\pm$1.1 & 36.8$\pm$6.6 & 43.4$\pm$6.1 & \textbf{54.7}$\pm${2.0} \\
    pen-human & 68.9  & 57.8$\pm$9.3 & 60.9$\pm$5.6 & 49.6$\pm$1.4 & \textbf{75.1}$\pm${1.5} \\
    pen-expert & 114.9 & 114.4$\pm$5.4 & 118.5$\pm$4.7 & 121.8$\pm$1.2 & \textbf{127.6}$\pm${2.0} \\
    door-cloned & 0.0 & 0.0$\pm$0.0 & 0.0$\pm$0.0 & 0.0$\pm$0.0 & \textbf{0.5}$\pm${0.2} \\
    door-human & 0.0 & -0.1$\pm$0.1 & 0.0$\pm$0.1 & 0.0$\pm$0.1 & \textbf{1.7}$\pm${0.1} \\
    door-expert & 99.0 & 104.2$\pm$0.3 & 103.7$\pm$0.2 & 102.5$\pm$1.2 & \textbf{105.3}$\pm${0.5} \\
    relocate-cloned & -0.3 & -0.3$\pm$0.0 & -0.3$\pm$0.0 & \textbf{-0.2}$\pm${0.0} & \textbf{-0.2}$\pm${0.0} \\
    relocate-human & \textbf{0.5} & 0.0$\pm$0.0 & 0.0$\pm$0.0 & 0.0$\pm$0.1 & 0.1$\pm$0.1 \\
    relocate-expert & 41.6 & 72.9$\pm$2.0 & 76.8$\pm$6.8 & 80.1$\pm$9.3 & \textbf{105.9}$\pm${1.0} \\
    hammer-cloned & 0.4 & 1.7$\pm$0.1 & 0.4$\pm$0.1 & 3.1$\pm$3.8 & \textbf{4.3}$\pm${1.6} \\
    hammer-human & 0.5 & 2.0$\pm$0.2 & 2.8$\pm$0.5 & 2.1$\pm$0.7 & \textbf{3.1}$\pm${2.2} \\
    hammer-expert & 107.2 & 126.8$\pm$1.0 & 126.9$\pm$1.0 & 126.8$\pm$1.3 & \textbf{128.9}$\pm${0.9} \\
    \midrule
    Total score & 476.7 & 520.6 & 526.5 & 529.2 & \textbf{607.0} \\
    \bottomrule
  \end{tabular}
 \vspace{-0.2cm}
\end{table}

\begin{table*}[t]
  \caption{Normalized average score comparison of CABI+TD3\_BC vs. baseline methods on the D4RL MuJoCo "-v0" dataset, where score 0 corresponds to a random policy performance and 100 corresponds to an expert policy performance. The results are averaged over the final 10 evaluations and 5 different random seeds. The highest mean scores are in \textbf{bold}.}
  \label{tab:mujocoscorecompare}
\vspace{0.2cm}
\renewcommand\tabcolsep{3pt}
  \renewcommand\arraystretch{1.05}
  \centering
  \footnotesize
  \begin{tabular}{@{}llllllllll@{}}
    \toprule
    Task Name & CABI+TD3\_BC & TD3\_BC & UWAC & MOPO & BCQ & BC & CQL & FisherBRC \\
    \midrule
    halfcheetah-random & 15.1$\pm$0.4 & 10.2 & 2.3 & \textbf{35.4} & 2.2 & 2.0 & 21.7 & 32.2 \\
    hopper-random & \textbf{11.9}$\pm$0.1 & 11.0 & 9.8 & 11.7 & 10.6 & 9.5 & 10.7 & 11.4 \\
    walker2d-random & 6.4$\pm$1.5 & 1.4 & 3.8 & \textbf{13.6} & 4.9 & 1.2 & 2.7 & 0.6 \\
    halfcheetah-medium-replay & 44.4$\pm$0.2 & 43.3 & 38.9 & \textbf{53.1} & 38.2 & 34.7 & 41.9 & 43.3 \\
    hopper-medium-replay & 31.3$\pm$0.7 & 31.4 & 18.0 & \textbf{67.5} & 33.1 & 19.7 & 28.6 & 35.6 \\
    walker2d-medium-replay & 29.4$\pm$1.3 & 25.2 & 8.4 & 39.0 & 15.0 & 8.3 & 15.8 & \textbf{42.6} \\
    halfcheetah-medium & \textbf{45.1}$\pm${0.1} & 42.8 & 37.4 & 42.3 & 40.7 & 36.6 & 37.2 & 41.3 \\
    hopper-medium & \textbf{100.4}$\pm${0.3} & 99.5 & 30.3 & 28.0 & 54.5 & 30.0 & 44.2 & 99.4 \\
    walker2d-medium & \textbf{82.0}$\pm${0.4} & 79.7 & 17.4 & 17.8 & 53.1 & 11.4 & 57.5 & 79.5 \\
    halfcheetah-medium-expert & \textbf{105.0}$\pm${0.2} & 97.9 & 40.6 & 63.3 & 64.7 & 67.6 & 27.1 & 96.1 \\
    hopper-medium-expert & \textbf{112.7}$\pm${0.0} & 112.2 & 95.4 & 23.7 & 110.9 & 89.6 & 111.4 & 90.6 \\
    walker2d-medium-expert & \textbf{108.4}$\pm${1.3} & 101.1 & 14.8 & 44.6 & 57.5 & 12.0 & 68.1 & 103.6 \\
    halfcheetah-expert & \textbf{107.6}$\pm$0.9 & 105.7 & 104.0 & - & 89.9 & 105.2 & 82.4 & 106.8 \\
    hopper-expert & \textbf{112.4}$\pm${0.1} & 112.2 & 109.1 & - & 107.0 & 111.5 & 111.2 & 112.3 \\
    walker2d-expert & \textbf{108.6}$\pm${1.5} & 105.7 & 88.4 & - & 102.3 & 56.0 & 103.8 & 79.9 \\
    \midrule
    Total Score & \textbf{1020.7} & 979.3 & 618.6 & - & 784.6 & 595.3 & 764.3 & 974.6\\
    \bottomrule
  \end{tabular}
\end{table*}

\noindent\textbf{CABI against forward/backward imagination.} We incorporate BCQ with the pure forward model, backward model, and CABI, and conduct extensive experiments on the Adroit tasks over 5 different random seeds. The forward model and reverse model are trained with the same configuration as CABI. The results are summarized in Table \ref{tab:wodoublecheckmechanism}. It can be seen that either the forward or reverse model results in limited improvement, which is consistent with the results of BOMI. As previously discussed, the forward model and reverse model may generate unreliable transitions. We see such evidence as the performance of BCQ falls on some of the tasks (e.g., \emph{pen-cloned}) if trained on mere forward or reverse imagination. BCQ+CABI, instead, outperforms BCQ+Forward and BCQ+Backward on all tasks. Hence, we conclude that CABI guarantees trustworthy transitions for training, and brings improvement on almost all of the tasks.

\noindent\textbf{CABI against other augmentation methods.} We further compare CABI against three data augmentation methods: (1) CABI-random where we replace the CVAE with the random policy as the rollout policy in CABI; (2) R-20 where we \emph{randomly} select 20\% synthetic transitions for bidirectional imagination; (3) EV-20 where we select 20\% transitions with the smallest \emph{ensemble variance} for bidirectional imagination, i.e., we evaluate the variance of the output of the ensemble of the forward and backward dynamics models and reject those with large variance. We use BCQ as the base algorithm and run experiments on four Adroit tasks for these augmentation methods with identical parameter setup as CABI (e.g., real data ratio). The results in Table \ref{tab:otheraugmentation} show that CABI performs consistently better than these methods. Since the data augmentation process of CABI is isolated from the policy optimization, we cannot leverage a random rollout policy because the generated actions of a random policy may possibly lie out of the span of the dataset, which can negatively affect the performance of the agent. Hence, CVAE is critical to ensure a safe and conservative data augmentation. Meanwhile, relying on the ensemble variance for data selection is not trustworthy as the models in the ensemble are trained on the identical data and may all incur wrong predictions but small variance.
\begin{table}[!h]
\caption{Normalized average score comparison on four Adroit tasks. The results are averaged over the final 10 evaluations and 5 different random seeds. CABI-random denotes the rollout policy in CABI is a random policy. R-20 denotes \textbf{R}andomly keep 20\% transitions, EV-20 denotes keep 20\% samples that have the smallest \textbf{E}nsemble \textbf{V}ariance.}
\label{tab:otheraugmentation}
\centering
\begin{tabular}{llllll}
\toprule
Task Name & BCQ & +CABI & +R-20 & +EV-20 & +CABI-random \\
\midrule
pen-cloned & 44.0 & \textbf{54.7}$\pm$2.0 & 41.2$\pm$3.0 & 40.4$\pm$2.0 & 37.6$\pm$7.8 \\
pen-expert & 114.9 & \textbf{127.6}$\pm$2.0 & 112.6$\pm$5.6 & 118.8$\pm$2.5 & 106.3$\pm$3.7 \\
hammer-cloned & 0.4 & \textbf{4.3}$\pm$1.6 & 0.9$\pm$0.6 & 0.4$\pm$0.1 & 0.3$\pm$0.0 \\ 
hammer-expert & 107.2 & \textbf{128.9}$\pm$0.9 & 104.2$\pm$24.6 & 125.5$\pm$5.5 & 103.8$\pm$1.5 \\
\bottomrule
\end{tabular}
\end{table}

\subsection{Broad Results on MuJoCo Dataset}
\label{sec:mujoco}

To show the generality of our method, we integrate CABI with another recent model-free offline RL method, TD3\_BC \cite{Fujimoto2021AMA}, and conduct experiments on 15 MuJoCo datasets. We widely compare CABI+TD3\_BC against other recent model-free offline RL methods, such as FisherBRC \cite{Kostrikov2021OfflineRL}, UWAC \cite{Wu2021UncertaintyWA}, CQL \cite{Kumar2020ConservativeQF}, and model-based batch RL method, MOPO \cite{Yu2020MOPOMO}. We run CABI+TD3\_BC over 5 different random seeds. We also run UWAC using the official codebase on the MuJoCo datasets over 5 different random seeds. The results of TD3\_BC, BC, CQL, FisherBRC are taken directly from \cite{Fujimoto2021AMA}, and the results of other baseline methods are taken from \cite{Wu2021UncertaintyWA}. 

The experimental results in Table \ref{tab:mujocoscorecompare} reveal that our approach exceeds all baseline methods on 10 out of 15 datasets, and is the strongest in terms of the total score. On almost all of the tasks, we observe performance improvement with CABI over the base TD3\_BC algorithm. Unfortunately, with the existence of behavioral cloning term, the performance improvement upon TD3\_BC is limited. Still, the experimental results in Table \ref{tab:adroitscorecompare} and \ref{tab:mujocoscorecompare} show that CABI is a powerful data augmentation method and can boost the performance of the base model-free offline RL methods.

\section{Conclusion and Limitations}
In this paper, we follow human nature and propose to do \emph{double check} during synthetic transition generation to ensure that the imagined samples are conservative and accurate. We admit samples that the forward model and reverse model agree on. Our method, CABI, involves training bidirectional dynamics models and rollout policies and can be combined with \emph{any} off-the-shelf model-free offline RL algorithms. Extensive experiments on the D4RL benchmarks show that our method significantly boosts the performance of the base model-free offline RL method, and can achieve competitive or better performance against recent baseline methods. For future work, it is interesting to evaluate CABI in the online setting and investigate whether it can benefit model-based online RL as well.

The major limitation of our proposed method lies in the computation cost as we train bidirectional dynamics models and rollout policies. However, since CABI is isolated from policy optimization, we can enhance the dataset beforehand.

\begin{ack}
This work was supported in part by the Science and Technology Innovation
2030-Key Project under Grant 2021ZD0201404, in part by the NSF China under Grant 61872009. The authors would like to thank the anonymous reviewers for their valuable comments and advice.
\end{ack}

\small
\bibliographystyle{abbrv}
\bibliography{neurips_2022.bib}

\section*{Checklist}

\begin{enumerate}

\item For all authors...
\begin{enumerate}
  \item Do the main claims made in the abstract and introduction accurately reflect the paper's contributions and scope?
    \answerYes{}
  \item Did you describe the limitations of your work?
    \answerYes{}
  \item Did you discuss any potential negative societal impacts of your work?
    \answerNA{}
  \item Have you read the ethics review guidelines and ensured that your paper conforms to them?
    \answerYes{}
\end{enumerate}

\item If you are including theoretical results...
\begin{enumerate}
  \item Did you state the full set of assumptions of all theoretical results?
    \answerNA{}
        \item Did you include complete proofs of all theoretical results?
    \answerNA{}
\end{enumerate}

\item If you ran experiments...
\begin{enumerate}
  \item Did you include the code, data, and instructions needed to reproduce the main experimental results (either in the supplemental material or as a URL)?
    \answerYes{}
  \item Did you specify all the training details (e.g., data splits, hyperparameters, how they were chosen)?
    \answerYes{}
        \item Did you report error bars (e.g., with respect to the random seed after running experiments multiple times)?
    \answerYes{}
        \item Did you include the total amount of compute and the type of resources used (e.g., type of GPUs, internal cluster, or cloud provider)?
    \answerYes{}
\end{enumerate}

\item If you are using existing assets (e.g., code, data, models) or curating/releasing new assets...
\begin{enumerate}
  \item If your work uses existing assets, did you cite the creators?
    \answerNA{}
  \item Did you mention the license of the assets?
    \answerNA{}
  \item Did you include any new assets either in the supplemental material or as a URL?
    \answerNo{}
  \item Did you discuss whether and how consent was obtained from people whose data you're using/curating?
    \answerNA{}
  \item Did you discuss whether the data you are using/curating contains personally identifiable information or offensive content?
    \answerNA{}
\end{enumerate}

\item If you used crowdsourcing or conducted research with human subjects...
\begin{enumerate}
  \item Did you include the full text of instructions given to participants and screenshots, if applicable?
    \answerNA{}
  \item Did you describe any potential participant risks, with links to Institutional Review Board (IRB) approvals, if applicable?
    \answerNA{}
  \item Did you include the estimated hourly wage paid to participants and the total amount spent on participant compensation?
    \answerNA{}
\end{enumerate}

\end{enumerate}


\clearpage

\appendix

\section{Experimental Setup of Toy RiskWorld Task}
\label{sec:setupriskworld}

In this section, we give the detailed experimental setup of our toy RiskWorld task. RiskWorld is a 2-dimensional, continuous state space, continuous action space environment as shown in Figure \ref{fig:riskworld}. We suppose the central point of RiskWorld is $(0,0)$, and the permitted range of RiskWorld gives $D\coloneqq [-1.5,1.5]\times[-1.5,1.5]$, i.e., the length of RiskWorld is 3. The state information in RiskWorld is composed of the coordinates of the agent, i.e., $s=(x,y), x,y\in[-1.5,1.5]$. There is a dangerous area $D_2$ locating at the central point with radius 0.5, i.e., $D_2\coloneqq \{(x,y)|x^2+y^2\le 0.5^2\}$. The agent randomly starts at a point in $D_1\coloneqq \{(x,y)|(x+1.5)^2+(y+1.5)^2\le 1, x<0, y<0\}$, and can take actions $a\in[-0.5,0.5]$. There is also a high reward zone locating at $D_3 \coloneqq \{(x,y)|(x-1.5)^2+(y-1.5)^2\le 0.8^2,x<1.5,y<1.5\}$. The reward function $r(s,a)$ is defined in (\ref{eq:rewardriskworld}).
\begin{equation}
    \label{eq:rewardriskworld}
    r(s,a) = \begin{cases}
    -3, \quad \mathrm{if} \, s\in D_2\coloneqq \{(x,y)|x^2+y^2\le0.5^2\}, \\
    1, \quad\quad \mathrm{if}\, s\in D_3\coloneqq \{(x,y)|(x-1.5)^2+(y-1.5)^2\le 0.8^2,x<1.5,y<1.5\}, \\
    0, \quad\quad \mathrm{if}\, s\in D\backslash (D_2\cup D_3).
    \end{cases}
\end{equation}
The agent will receive a large minus reward of $-3$ if it steps into the dangerous zone $D_2$, and the done flag will turn into \textit{true}. The agent is not allowed to step out of the legal region $D$. The episode length for RiskWorld is set to be 300. The RiskWorld is intrinsically a sparse reward environment. We run a random policy on RiskWorld for $10^4$ timesteps and log the transition data it collected during interactions to form a static dataset \emph{RiskWorld-random}.

Model-based reinforcement learning (RL) learns either forward dynamics or reverse dynamics of the environment \cite{Sutton2005ReinforcementLA, goyal2018recall}, and can produce imaginary transitions for training, which has been widely demonstrated to be effective in improving the sample efficiency of RL in the online setting \cite{Buckman2018SampleEfficientRL, Janner2019WhenTT}. The forward dynamics model $\hat{p}_\psi(s^\prime,r|s,a)$ predicts the next state and the corresponding reward function given the current state and action, and the reverse dynamics model $\hat{p}_\phi(s,r|s^\prime,a)$ outputs the previous state and reward signal given action and the next state. Bidirectional modeling combines both forward dynamics model and backward dynamics model.

To compare different ways of imagination, i.e., the forward imagination, reverse imagination, and bidirectional imagination with the double check mechanism, we train a forward dynamics model, a backward dynamics model, and a bidirectional dynamics model on RiskWorld-random dataset, respectively. We represent the forward and reverse dynamics model by training a probabilistic neural network. The forward model $\hat{p}_\psi(s^\prime,r|s,a)$ parameterized by $\psi$ receives the current state and action as input, and outputs a multivariate Gaussian distribution that predicts the next state and reward as shown in (\ref{eq:forwarddynamics}).
\begin{equation}
    \label{eq:forwarddynamics}
    \hat{p}_\psi(s^\prime,r|s,a) = \mathcal{N}({\bf{\mu}}_\psi(s,a), {\bf{\Sigma}}_\psi(s,a)),
\end{equation}
where ${\bf\mu}_\psi$ and ${\bf\Sigma}_\psi$ represent the mean and variance of the forward model $\hat{p}_\psi(s^\prime,r|s,a)$, respectively.

Similarly, for the backward model $\hat{p}_\phi(s,a|s^\prime,a)$ parameterized by $\phi$, it adopts the next state and action as input and outputs a multivariate normal distribution predicting reward signal and the previous state (see (\ref{eq:backwarddynamics})).
\begin{equation}
    \label{eq:backwarddynamics}
    \hat{p}_\phi(s,a|s^\prime,a) = \mathcal{N}({\bf\mu}_\phi(s^\prime,a),{\bf\Sigma}_\phi(s^\prime,a)),
\end{equation}
where ${\bf\mu}_\phi$ and ${\bf\Sigma}_\phi$ denote the mean and variance of the backward model $\hat{p}_\phi(s,r|s^\prime,a)$, respectively.

The probabilistic neural network is modeled by a multi-layer neural network that consists of 4 feedforward layers with 400 hidden units. We adopt \emph{swish} activation for each intermediate layer. Following prior works \cite{Janner2019WhenTT, Yu2020MOPOMO, Kidambi2020MOReLM}, we train an ensemble of seven such probability neural networks for both the forward and backward model. We use a hold-out set made up of 1000 transitions to validate the performance of the trained dynamics, and select the five models that have the best performance accordingly. When performing forward or reverse imagination, we randomly pick one model out of the five best model candidates to generate synthetic trajectories per step. Considering the simplicity of the toy RiskWorld task, we train both the forward dynamics model and reverse dynamics model for 100 epochs, and the rollout length (horizon) is set to be 3 for the forward model, reverse model, and bidirectional model. We use the trained dynamics model (forward, reverse, bidirectional) to generate $10^4$ imaginary transition samples, and log their model buffer respectively. We plot in Figure \ref{fig:toyexample} the model buffer of these dynamics models and the raw static dataset obtained by running the random policy.

\section{Datasets and Evaluation Setting on the D4RL Benchmarks}
In this section, we give a detailed description of the datasets we used in this paper, and also describe the evaluation setting that is adopted on the D4RL benchmarks \cite{Fu2020D4RLDF}. D4RL is specially designed for evaluating offline RL (or batch RL) algorithms, which covers the dimensions that offline RL may encounter in practical applications, such as passively logged data, human demonstrations, etc.

\subsection{Adroit datasets and MuJoCo datasets}
\begin{figure}[!htb]
    \centering
    \subfigure[Pen]{
    \label{fig:pen}
    \includegraphics[scale=0.3]{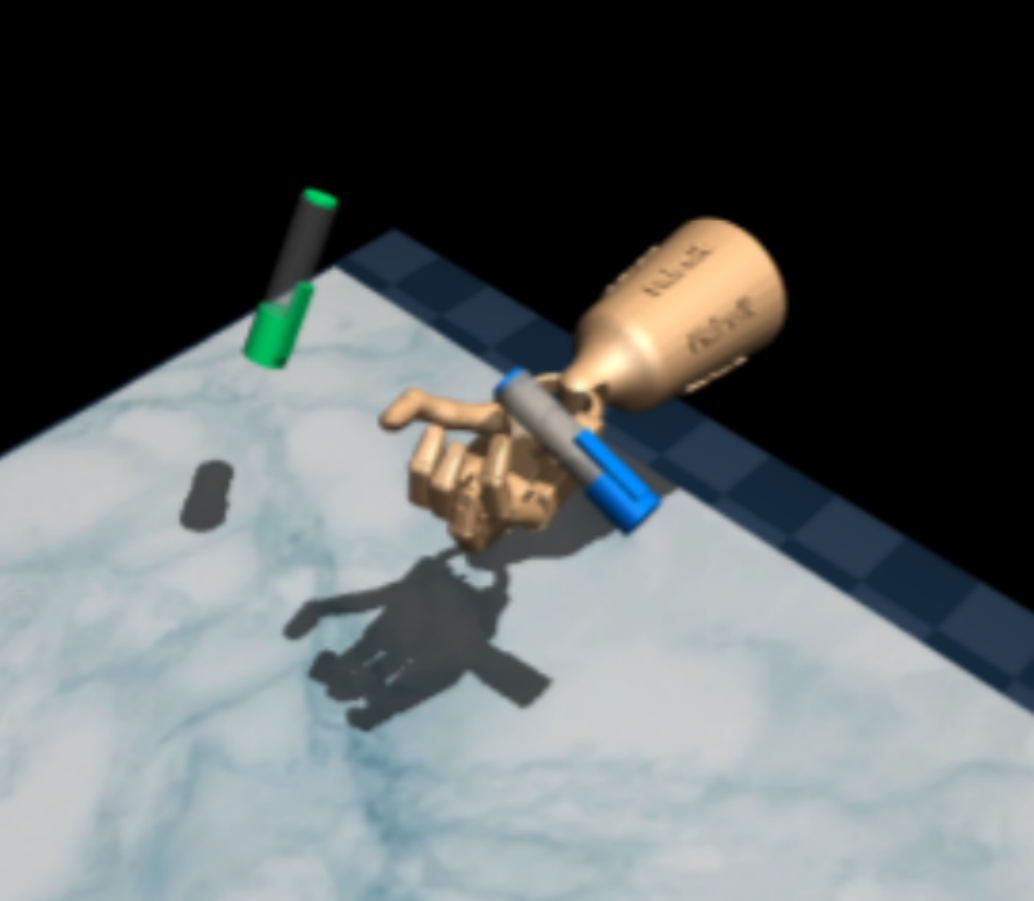}
    }\hspace{-2mm}
    \subfigure[Door]{
    \label{fig:door}
    \includegraphics[scale=0.3]{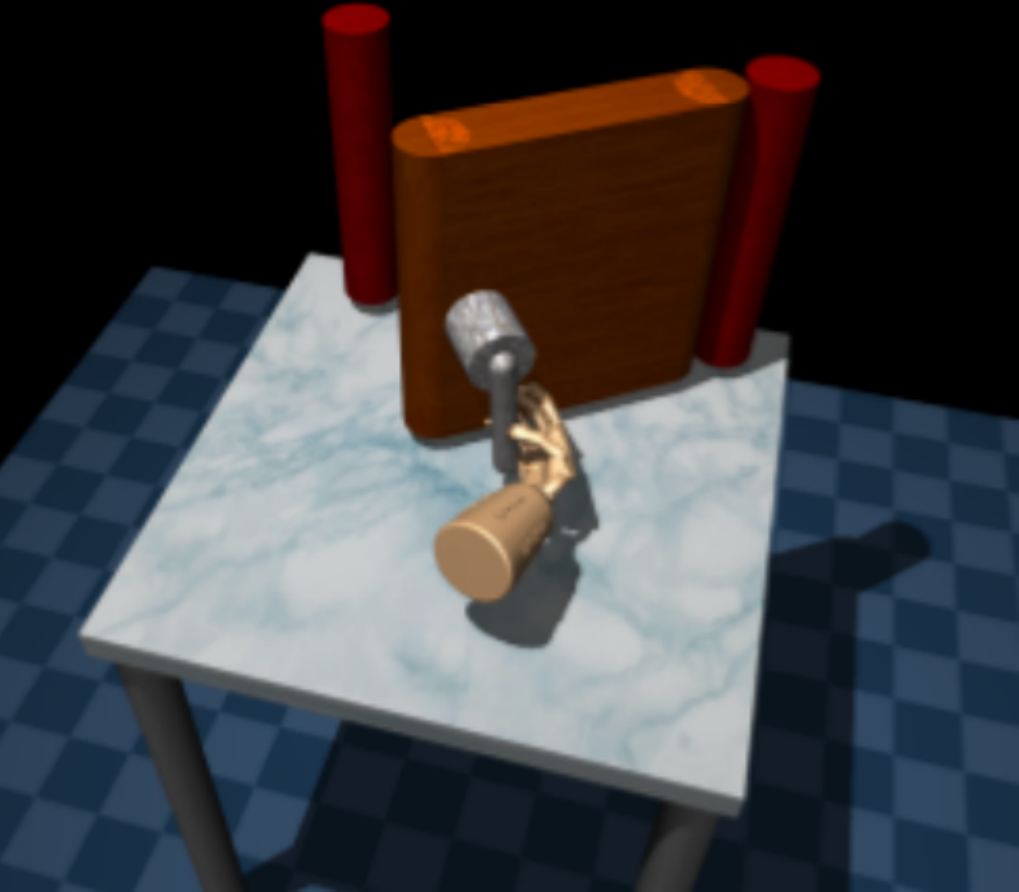}
    }\hspace{-2mm}
    \subfigure[Relocate]{
    \label{fig:relocate}
    \includegraphics[scale=0.3]{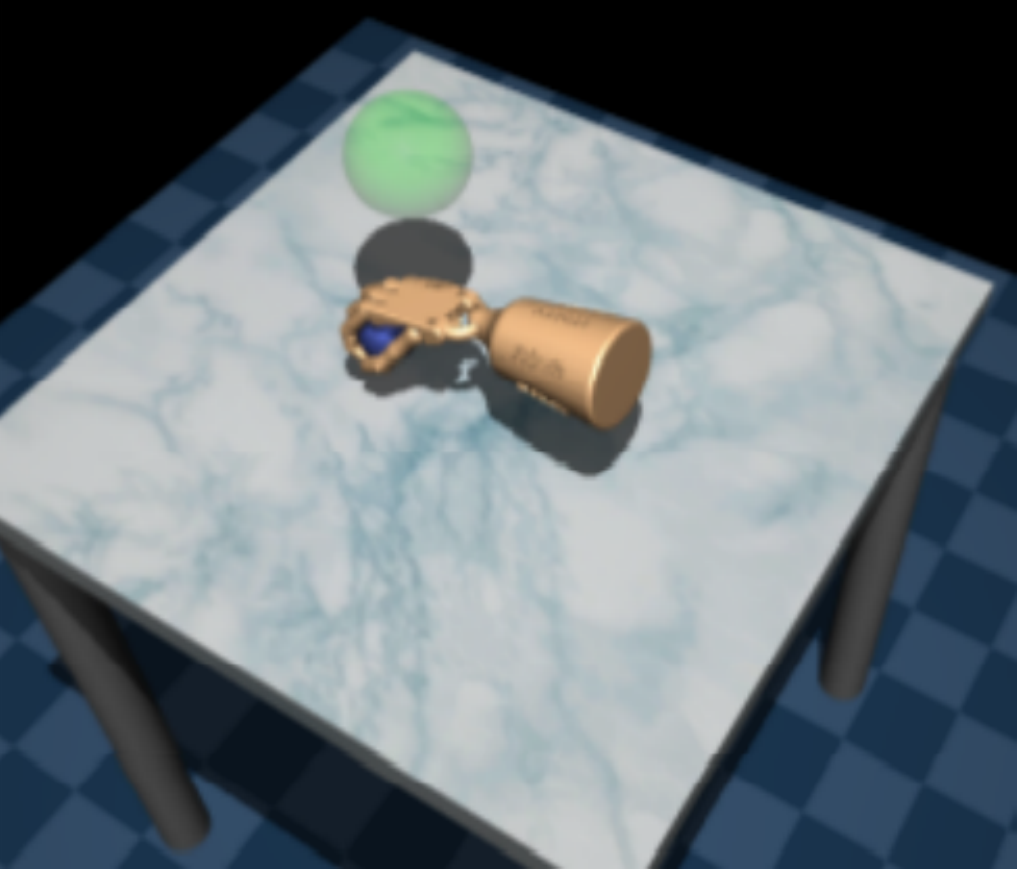}
    }\hspace{-2mm}
    \subfigure[Hammer]{
    \label{fig:hammer}
    \includegraphics[scale=0.3]{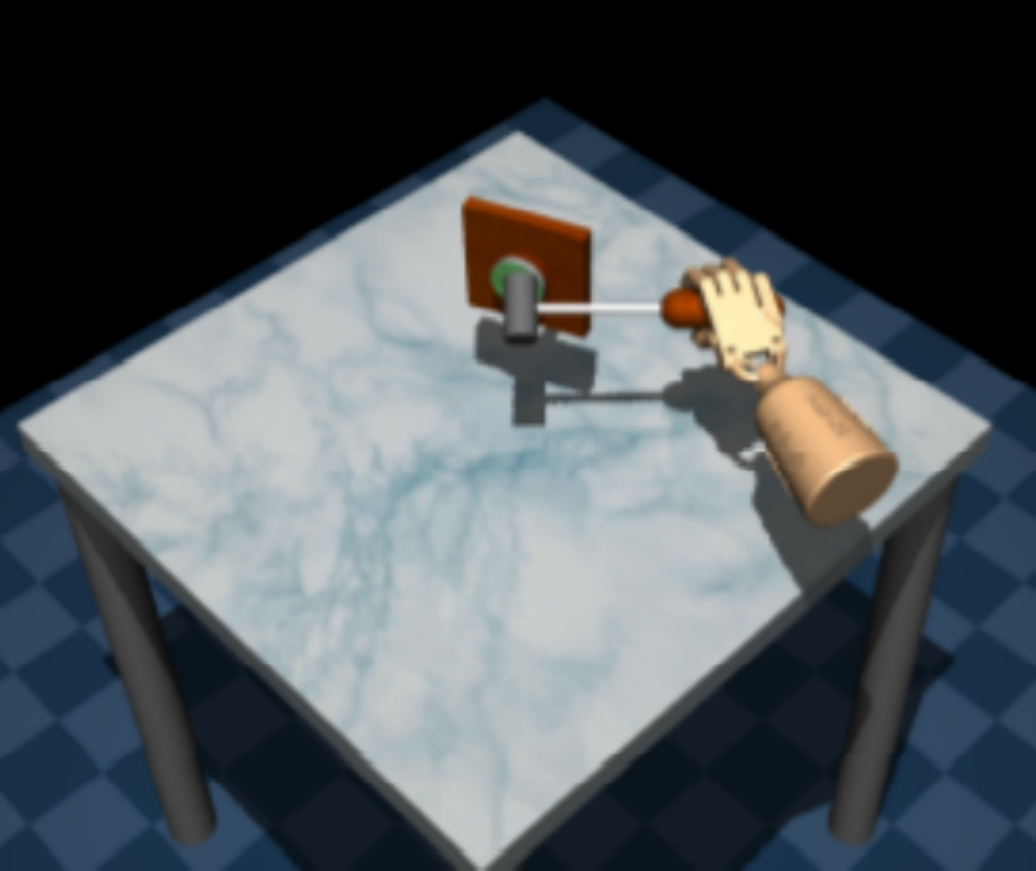}
    }\hspace{-2mm}
    \caption{Adroit datasets. There are four tasks, namely pen, door, relocate, and hammer.}
    \label{fig:adroitviaual}
\end{figure}

The Adroit dataset involves controlling a 24-DoF simulated Shadow Hand robot to perform tasks like hammering a nail, opening a door, twirling a pen, and picking/moving a ball, as shown in Figure \ref{fig:adroitviaual}. The Adroit domain is super challenging for even online RL algorithms because: (1) the dataset contains narrow human demonstrations; (2) this domain solves sparse reward, high-dimensional robotic manipulation tasks. There are four tasks in the dataset, and there are three types of datasets for each task, \emph{cloned, human, expert}. \textbf{human:} a small number of demonstrations operated by a human (25 trajectories per task). \textbf{expert:} a large amount of expert data from a fine-tuned RL policy. \textbf{cloned:} a large amount of data generated by performing imitation learning on the human demonstrations, running the policy, and mixing the data at a 50-50 ratio with the demonstrations. Dataset mixing is involved for \emph{cloned} as the cloned policies themselves fail on the tasks, making the dataset otherwise hard to learn from.

\begin{figure}[!htb]
    \centering
    \subfigure[HalfCheetah]{
    \label{fig:halfcheetah}
    \includegraphics[scale=0.295]{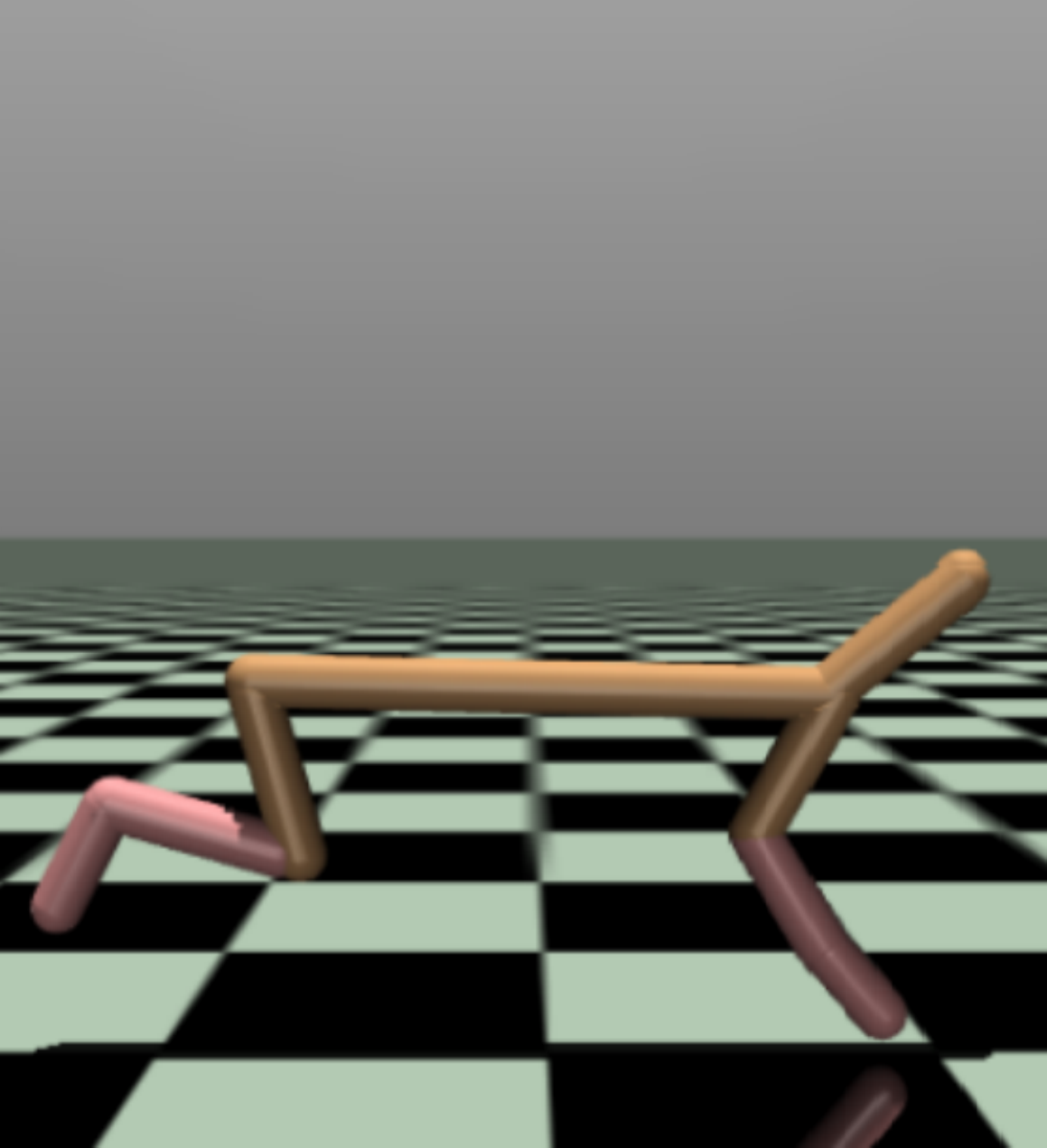}
    }
    \subfigure[Hopper]{
    \label{fig:hopper}
    \includegraphics[scale=0.3]{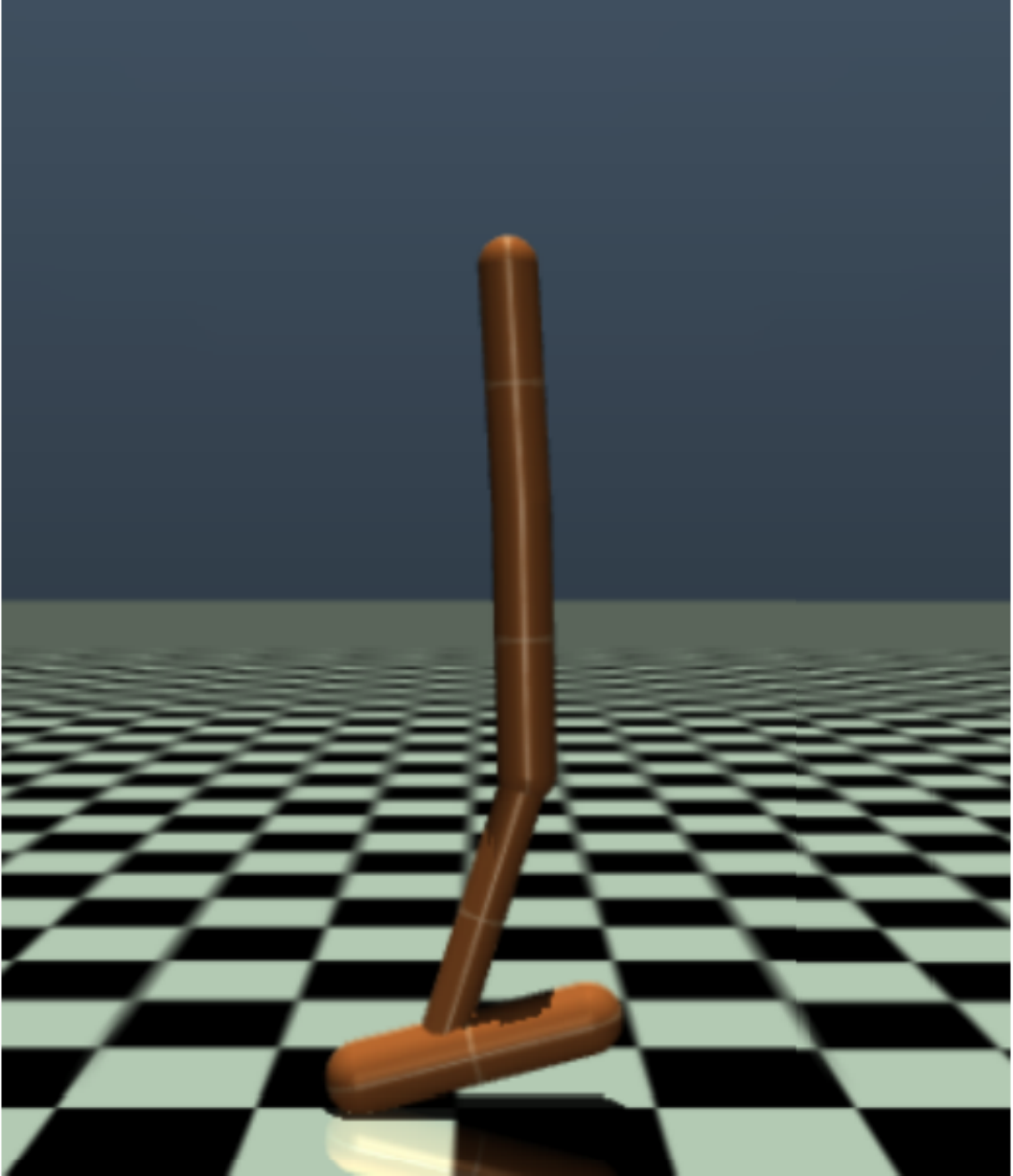}
    }
    \subfigure[Walker2d]{
    \label{fig:walker2d}
    \includegraphics[scale=0.3]{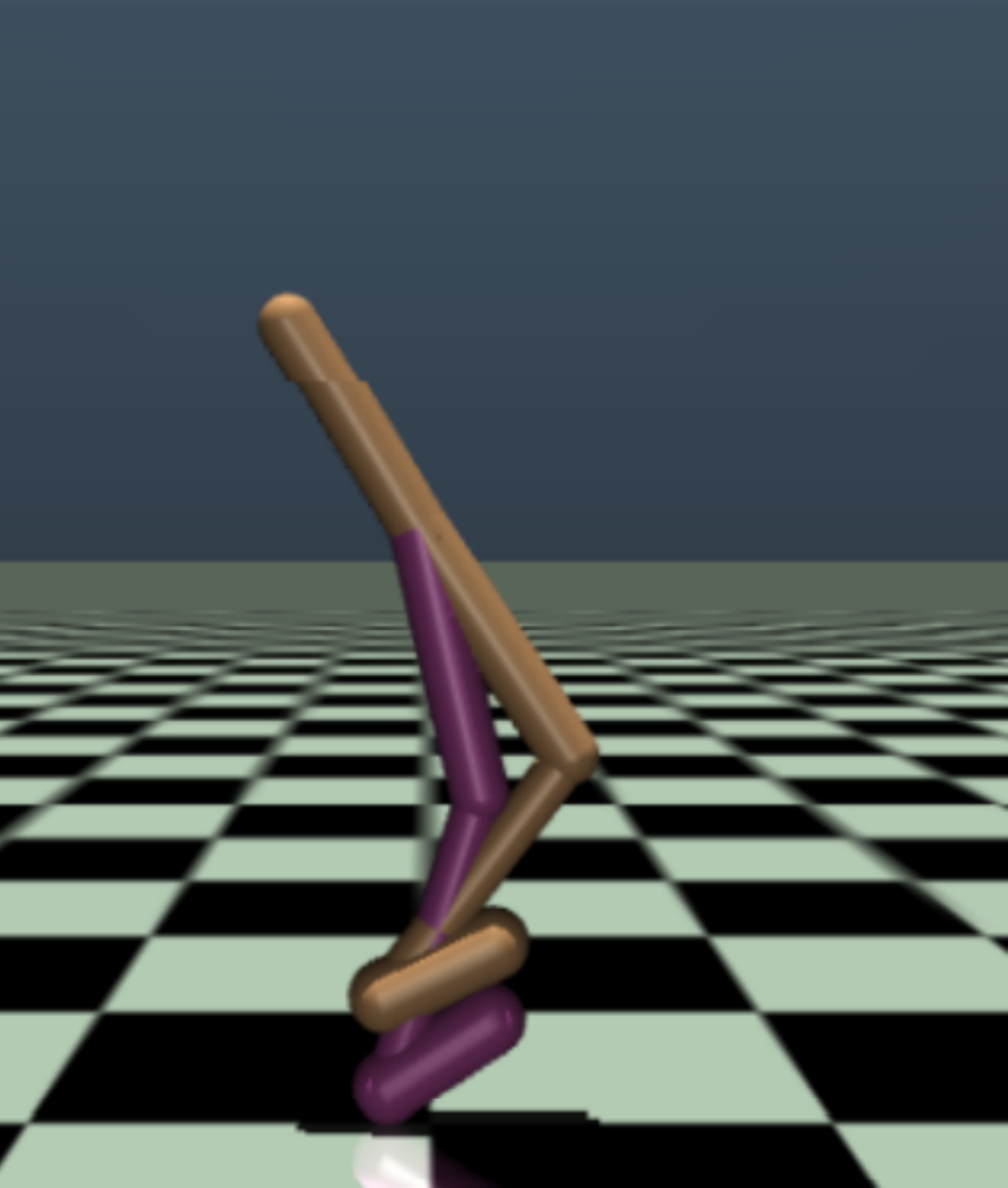}
    }
    \caption{MuJoCo datasets. We conduct experiments on halfcheetah, hopper, and walker2d tasks.}
    \label{fig:mujocoviaual}
\end{figure}

The MuJoCo dataset is collected during the interactions with the continuous action environments in Gym \cite{Brockman2016OpenAIG} simulated by MuJoCo \cite{Todorov2012MuJoCoAP}. We adopt three tasks in this dataset, \emph{halfcheetah, hopper, walker2d} as illustrated in Figure \ref{fig:mujocoviaual}. Each task in the MuJoCo dataset contains five types of datasets, \emph{random, medium, medium-replay, medium-expert, expert}. \textbf{random:} a large amount of data from a random policy. \textbf{medium:} experiences collected from an early-stopped SAC policy for 1M steps. \textbf{medium-replay:} replay buffer of a policy trained up to the performance of the medium agent. \textbf{expert:} a large amount of data gathered by the SAC policy that is trained to completion. \textbf{medium-expert:} a large amount of data by mixing the medium data and expert data at a 50-50 ratio.

Note that the Adroit dataset is qualitatively different from the MuJoCo Gym dataset because (1) there do not exist human demonstrations in the MuJoCo dataset; (2) the reward is dense in MuJoCo, making it less challenging to learn from; (3) the dimension of transitions in MuJoCo is low compared with Adroit. It is hard for even online RL algorithms to learn useful policies on the Adroit tasks, while it is easy for online RL methods to achieve superior performance on the MuJoCo environments.

\subsection{Evaluation setting in D4RL}
D4RL suggests using the normalized score metric to evaluate the performance of the offline RL algorithms \cite{Fu2020D4RLDF}. Denote the expected return of a random policy on the dataset as $C_r$ (reference min score), and the expected return of an expert policy as $C_e$ (reference max score). Suppose that an offline RL algorithm achieves an expected return of $C$ after training on the given dataset. Then the normalized score $\tilde{C}$ is given by (\ref{eq:normalizedscore}).
\begin{equation}
    \label{eq:normalizedscore}
    \tilde{C} = \frac{C - C_r}{C_e - C_r}\times 100 = \frac{C - \mathrm{performance}\ \mathrm{of}\ \mathrm{random}\ \mathrm{policy}}{\mathrm{performance}\ \mathrm{of}\ \mathrm{expert}\ \mathrm{policy} - \mathrm{performance}\ \mathrm{of}\ \mathrm{random}\ \mathrm{policy}} \times 100.
\end{equation}

The normalized score ranges roughly from 0 to 100, where 0 corresponds to the performance of a random policy and 100 corresponds to the performance of an expert policy. We give the detailed reference min score $C_r$ and reference max score $C_e$ in Table \ref{tab:referencescore}, where all of the tasks share the same reference min score and reference max score across different types of datasets.

\begin{table}[!htb]
  \caption{The referenced min score and max score for the Adroit dataset and MuJoCo dataset in D4RL.}
  \label{tab:referencescore}
  \vspace{0.2cm}
  \centering
  \begin{tabular}{llcc}
    \toprule
    Domain & Task Name   & Reference min score $C_r$ & Reference max score $C_e$ \\
    \midrule
    Adroit & pen  & 96.26 & 3076.83  \\
    Adroit & door & $-$56.51 & 2880.57 \\
    Adroit & relocate & $-$6.43 & 4233.88 \\
    Adroit & hammer & $-$274.86 & 12794.13 \\
    \midrule
    MuJoCo & halfcheetah & $-$280.18 & 12135.0 \\
    MuJoCo & hopper & $-$20.27 & 3234.3 \\
    MuJoCo & Walker2d & 1.63 & 4592.3 \\
    \bottomrule
  \end{tabular}
\end{table}

\section{Implementation Details and Hyperparameters}
\label{sec:implementationdetails}
\subsection{Implementation details}
In this section, we give implementation details and hyperparameters for Confidence-Aware Bidirectional Offline Model-Based Imagination (CABI). We represent the approximated forward dynamics and reward model, and backward dynamics and reward model by training a probabilistic neural network. The configuration of the probabilistic neural network is identical to Appendix \ref{sec:setupriskworld}. That is, the forward model and reverse model are modeled as a multivariate Gaussian distribution with mean ${\bf \mu}$ and variance ${\bf \Sigma}$. For the forward model $\hat{p}_\psi(s^\prime,r|s,a)$ parameterized by $\psi$, it accepts the current state and action as input and generates the next state and reward. The backward model $\hat{p}_\phi(s,r|s^\prime,a)$ parameterized by $\phi$ receives the next state and current action as input and outputs the former state and scalar reward. The probabilistic neural network is modeled by a multi-layer neural network that contains 4 feedforward layers, with 400 hidden units in each layer, and a \emph{swish} activation in each intermediate layer. We train an ensemble of seven such probabilistic neural networks and select the best five models based on their performance on a hold-out set made up of 1000 transitions from the offline dataset.

As a data augmentation method, CABI does not actively generate actions during the training process, i.e., the policy optimization process is isolated from the data generation process. We use a conditional variational autoencoder (CVAE) to approximate the behavior policy in the static dataset. We give brief introduction to the VAE in Appendix \ref{sec:vaebackground}. The CVAE ${G}$ contains an encoder ${E}$ and a decoder ${D}$. Both the encoder and the decoder in the forward rollout policy and backward rollout policy contain two intermediate layers with 750 hidden units each layer. We adopt \emph{relu} activation for each intermediate layer. Specifically, we train a forward rollout policy with a CVAE ${G}_\theta^{\mathrm{fwd}}(s)$, which contains an encoder ${E}_{\xi_1}(s,a)$ and a decoder ${D}_{\nu_1}(s,z)$, $\theta = \{\xi_1,\nu_1\}$, and a reverse rollout policy ${G}_\omega^{\mathrm{bwd}}(s^\prime) = \{ {E}_{\xi_2}(s^\prime,a), {D}_{\nu_2}(s^\prime,z) \}$, where $\omega = \{\xi_2,\nu_2\}$. Note that the forward rollout policy ${G}_\theta^{\mathrm{fwd}}(s)$ and reverse rollout policy ${G}_\omega^{\mathrm{bwd}}(s^\prime)$ sample actions using stochastic inference from an underlying latent space, so as to increase diversity in the generated actions.

Intrinsically, CABI can be combined with \emph{any} off-the-shelf model-free offline RL algorithms. In this work, we incorporate CABI with BCQ \cite{Fujimoto2019OffPolicyDR} and TD3\_BC \cite{Fujimoto2021AMA}, and conduct extensive experiments on the Adroit dataset and MuJoCo dataset on the D4RL benchmarks, respectively. There are generally three steps when combining CABI with model-free offline RL methods: (1) \textbf{Model training.} We first train bidirectional dynamics models and bidirectional rollout policies using the raw static offline dataset $\mathcal{D}_{\mathrm{env}}$; (2) \textbf{Data generation.} After the bidirectional models and rollout policies are well trained, we utilize them to generate imaginary trajectories, while conducting double check and admitting high-confidence transitions simultaneously. This will induce the model generated dataset $\mathcal{D}_{\mathrm{model}}$; (3) \textbf{Policy optimization.} We then merge the real dataset $\mathcal{D}_{\mathrm{env}}$ with the imagined dataset $\mathcal{D}_{\mathrm{model}}$ to form a composite dataset $\mathcal{D}_{\mathrm{total}}$, i.e., $\mathcal{D}_{\mathrm{total}} = \mathcal{D}_{\mathrm{env}}\cup \mathcal{D}_{\mathrm{model}}$. The mini-batch samples used for training the model-free offline RL algorithms come from $\mathcal{D}_{\mathrm{env}}$ and $\mathcal{D}_{\mathrm{model}}$. To be specific, we define the ratio of data come from $\mathcal{D}_{\mathrm{env}}$ (real data) as $\eta$. Suppose we use a mini-batch size of $N$ for training the algorithm. Then for each optimization step, we sample $\eta N$ samples from $\mathcal{D}_{\mathrm{env}}$, and sample $(1-\eta)N$ samples from $\mathcal{D}_{\mathrm{model}}$. We pick the optimal real data ratio $\eta$ among $\{0.1,0.3,0.5,0.7,0.9\}$ using grid search.

\noindent\textbf{Influence of $k$.} For the double check mechanism, we keep the top 20\% of samples in a mini-batch that the forward model and backward model have the smallest disagreements. As explained in Section \ref{sec:dataaugmentation}, we do not set a threshold and then admit samples that the deviation of the forward model and backward model on them are smaller than the threshold, because it requires human knowledge and the threshold is task-specific. However, in real-world problems, we cannot always have full knowledge about the system. For better flexibility and generality, we choose to keep top $k$\% samples. Note that if we use $k=0$, then the influence of data augmentation will be excluded. If we use $k=100$, then CABI will degenerate into Bidirectional Model-based Offline Imagination (BOMI), where there is no double check procedure. Intuitively, if $k$ is small, few samples can be left, which may negatively affect the advantages of data augmentation with a model. While if $k$ is large, some poorly imagined transitions may be included in the model buffer. We simply set $k=20$ by default, and use it throughout our experiments on the Adroit dataset and MuJoCo dataset. We conduct experiments on two types of MuJoCo datasets, \emph{random, medium}, with varied $k$ in $\{0, 10, 20, 50, 100\}$ over 5 different random seeds. The results are shown in Table \ref{tab:topkablation}, where we observe performance drop for both large $k$ and small $k$. We hence set $k=20$ for all of our experiments.

\begin{table}[!htb]
  \caption{Normalized average score of CABI+TD3\_BC on two datasets, random and medium, from MuJoCo dataset with different values of $k$. CABI+TD3\_BC with $k=0$ degenerates into vanilla TD3\_BC, and CABI+TD3\_BC with $k=100$ turns into BOMI+TD3\_BC. We \textbf{bold} the highest mean.}
  \label{tab:topkablation}
    \vspace{0.2cm}
  \centering
  \begin{tabular}{lccccc}
    \toprule
    Task Name & $k=0$ & $k=10$ & $k=20$ & $k=50$ & $k=100$ \\
    \midrule
    halfcheetah-random & 10.2 & 14.8 & \textbf{15.1} & 14.0 & 11.4 \\
    hopper-random & 11.0 & 10.3 & \textbf{11.9} & 10.6 & 9.3 \\
    walker2d-random & 1.4 & 4.9 & \textbf{6.4} & 5.8 & 4.3 \\
    halfcheetah-medium & 42.8 & 44.6 & \textbf{45.1} & 44.4 & 44.3 \\
    hopper-medium & 99.5 & 99.7 & \textbf{100.4} & 32.3 & 3.1 \\
    walker2d-medium & 79.7 & 78.7 & \textbf{82.0} & 79.8 & 78.6 \\
    \bottomrule
  \end{tabular}
\end{table}

\noindent\textbf{Computation time and compute infrastructure.} The computation time for CABI ranges from 4 to 14 hours on the MuJoCo and Adroit tasks. The model training time differs on different types of datasets, e.g., it takes much less time to train our bidirectional models and rollout policies on \emph{medium-replay} and \emph{human} datasets (about 40 minutes), while it takes comparatively longer time to train on other types of datasets (about 2-6 hours). TD3\_BC consumes about 3 hours to run on all MuJoCo datasets, and BCQ takes about 6-8 hours to train on the Adroit tasks. We run both CABI+BCQ and CABI+TD3\_BC for $1\times 10^6$ timesteps. We additionally run IQL \cite{kostrikov2022offline}, and the results are presented in Section \ref{sec:iql}. IQL takes about 3-7 hours to run on all tasks with $1\times 10^6$ timesteps. We give detailed compute infrastructure in Appendix \ref{sec:computinginfrastructure}.

\noindent\textbf{Discussion on CABI and ROMI \cite{Wang2021OfflineRL}.} A recent work, Reverse Offline Model-based Imagination (ROMI) \cite{Wang2021OfflineRL}, explores the data augmentation in offline RL via training a reverse dynamics model. It is worth noting that we do not directly compare with ROMI+BCQ as there are \textit{many secondary components} in the codebase of ROMI (\href{https://github.com/wenzhe-li/romi}{https://github.com/wenzhe-li/romi}), e.g., prioritized experience replay, modifying state information, adopting varied rollout policies for different domains, etc. CABI represents the rollout policy using only conditional variational autoencoder (CVAE). Also, ROMI assumes that the termination functions are known. However, CABI does not include any prior knowledge about termination conditions, even on simple MuJoCo tasks. That generally follows the key claims in the recent work of \cite{Fujimoto2021AMA}. For a fair comparison, we disable the forward model as well as the double check mechanism in CABI to get our reverse model. As for the comparison with forward imagination, we disable the double check mechanism and the reverse dynamics part of CABI to get our pure forward model.

\noindent\textbf{Ensemble variance.} In the main text, we compare our CABI against EV-20, which rejects transitions with large ensemble variance. The ensemble variance is calculated in the following way where we take the forward imagination as an example: suppose we have an ensemble of forward dynamics models $f_i(\cdot|s,a),i=1,\ldots,N$ with the ensemble size $N$. Then each model in the ensemble predicts a next state $s_i^\prime$. We then have a collection of next state $\{s_i^\prime\},i=1,\ldots,N$. We then randomly pick one next state while recording the variance in the ensemble at the same time. We reject the generated next state if the variance in the ensemble is large. That is, we evaluate the variance of $\{s_i^\prime\},i=1,\ldots,N$. We sort the transitions in a batch by their calculated variance, and only trust the top 20\% that have the smallest ensemble variance.

\noindent\textbf{Other implementation details.} On the Adroit tasks, we combine CABI with BCQ, and compare against recent state-of-the-art methods, including vanilla BCQ \cite{Fujimoto2019OffPolicyDR}, UWAC \cite{Wu2021UncertaintyWA}, CQL \cite{Kumar2020ConservativeQF}, MOPO \cite{Yu2020MOPOMO}, and COMBO \cite{Yu2021COMBOCO}, etc. On the MuJoCo domain, we incorporate CABI with TD3\_BC \cite{Fujimoto2021AMA}, and compare with baseline methods like FisherBRC \cite{Kostrikov2021OfflineRL}, UWAC, CQL, BEAR \cite{Kumar2019StabilizingOQ}, MOPO, etc. Note that we omit some baseline methods, such as AWAC \cite{Nair2020AcceleratingOR} and BRAC \cite{Wu2019BehaviorRO} in the MuJoCo domain, as they do not obtain good enough performance for comparison. We run UWAC with the official codebase (\href{https://github.com/apple/ml-uwac}{https://github.com/apple/ml-uwac}), and so is MOPO (\href{https://github.com/tianheyu927/mopo}{https://github.com/tianheyu927/mopo}). We run MOPO on the Adroit tasks as those results are not reported in the original paper, and we take the results of MOPO on the MuJoCo datasets from \cite{Wu2021UncertaintyWA} directly. We re-run UWAC on the Adroit domain and MuJoCo domain because, unfortunately, we cannot reproduce the results reported in its original paper. All baseline methods are run for $1\times 10^6$ timesteps over 5 different random seeds.

\subsection{Hyperparameters}
In this subsection, we give the detailed hyperparameter setup for our experiments in Table \ref{tab:hyperparameter}. We keep the top 20\% samples in a mini-batch for all tasks. For simplicity, we use identical rollout length for the forward model and backward model. For all of the MuJoCo tasks and most of the Adroit tasks, the rollout length for both the forward model and backward model is set to be 3, which yields a total horizon of 6. On datasets that the model disagreement are comparatively large for long horizons (e.g., \emph{pen-expert}, see Table \ref{tab:modeldisagreementcabi}), we set the forward and backward horizon as 1, which leads to a total horizon of 2. We use the forward and backward horizon 5 for \emph{hammer-human} as we experimentally find that it performs better. On other tasks, the forward and backward horizon is set to be 3 by default. Note that the model disagreement on MuJoCo datasets are smaller than 0.1 for all horizons, and the trained forward and backward model well fits MuJoCo datasets. We therefore adopt the forward and backward horizon of 3 for all of these tasks.

\begin{table}[!htb]
  \caption{Hyperparameters setup in our experiments with CABI on the Adroit dataset and MuJoCo dataset. ForH = Forward Horizon, BackH = Backward Horizon.}
  \label{tab:hyperparameter}
    \vspace{0.2cm}
  \centering
  \begin{tabular}{lllccc}
    \toprule
    Domain & Dataset Type & Task Name   & ForH & BackH & Real Data Ratio $\eta$ \\
    \midrule
    Adroit & human & pen  & 1 & 1 & 0.7 (BCQ), 0.5 (IQL) \\
    Adroit & human & door & 3 & 3 & 0.7 (BCQ), 0.5 (IQL) \\
    Adroit & human & relocate & 3 & 3 & 0.9 (BCQ), 0.5 (IQL)  \\
    Adroit & human & hammer & 5 & 5 & 0.5 (BCQ), 0.7 (IQL)  \\
    Adroit & cloned & pen  & 1 & 1 & 0.5 (BCQ), 0.5 (IQL)  \\
    Adroit & cloned & door  & 3 & 3 & 0.5 (BCQ), 0.7 (IQL) \\
    Adroit & cloned & relocate  & 3 & 3 & 0.3 (BCQ), 0.5 (IQL)  \\
    Adroit & cloned & hammer  & 3 & 3 & 0.5 (BCQ), 0.7 (IQL) \\
    Adroit & expert & pen  & 1 & 1 & 0.7 (BCQ), 0.9 (IQL)  \\
    Adroit & expert & door  & 3 & 3 & 0.7 (BCQ), 0.9 (IQL) \\
    Adroit & expert & relocate  & 3 & 3 & 0.9 (BCQ), 0.7 (IQL) \\
    Adroit & expert & hammer  & 1 & 1 & 0.9 (BCQ), 0.5 (IQL) \\
    \midrule
    MuJoCo & random & halfcheetah & 3 & 3 & 0.7 (TD3\_BC), 0.7 (IQL) \\
    MuJoCo & random & hopper & 3 & 3 & 0.1 (TD3\_BC), 0.7 (IQL) \\
    MuJoCo & random & walker2d & 3 & 3 & 0.1 (TD3\_BC), 0.7 (IQL) \\
    MuJoCo & medium & halfcheetah & 3 & 3 & 0.7 (TD3\_BC), 0.7 (IQL) \\
    MuJoCo & medium & hopper & 3 & 3 & 0.9 (TD3\_BC), 0.7 (IQL) \\
    MuJoCo & medium & walker2d & 3 & 3 & 0.7 (TD3\_BC), 0.9 (IQL) \\
    MuJoCo & medium-replay & halfcheetah & 3 & 3 & 0.5 (TD3\_BC), 0.7 (IQL) \\
    MuJoCo & medium-replay & hopper & 3 & 3 & 0.7 (TD3\_BC), 0.7 (IQL) \\
    MuJoCo & medium-replay & walker2d & 3 & 3 & 0.5 (TD3\_BC), 0.9 (IQL) \\
    MuJoCo & medium-expert & halfcheetah & 3 & 3 & 0.7 (TD3\_BC), 0.9 (IQL) \\
    MuJoCo & medium-expert & hopper & 3 & 3 & 0.9 (TD3\_BC), 0.9 (IQL) \\
    MuJoCo & medium-expert & walker2d & 3 & 3 & 0.7 (TD3\_BC), 0.9 (IQL) \\
    MuJoCo & expert & halfcheetah & 3 & 3 & 0.7 (TD3\_BC), 0.9 (IQL) \\
    MuJoCo & expert & hopper & 3 & 3 & 0.9 (TD3\_BC), 0.9 (IQL) \\
    MuJoCo & expert & walker2d & 3 & 3 & 0.7 (TD3\_BC), 0.9 (IQL)  \\
    \bottomrule
  \end{tabular}
\end{table}

We search for the best $\eta$ over $\{0.1,0.3,0.5,0.7,0.9\}$. We find that the real data ratio $\eta=0.7$ and $\eta=0.9$ are generally effective for CABI. The best ratio $\eta$ strongly depends on the dataset and may need to be tuned manually. For example, \emph{random} dataset in the MuJoCo domain and \emph{cloned} dataset in the Adroit domain are poor for training naturally, and small $\eta$ is therefore needed. While for \emph{expert} dataset or \emph{medium} dataset, a comparatively large $\eta$ is better. 

\section{Model Prediction Error and Model Disagreement}
In this section, we are interested in exploring (1) can CABI generate more trustworthy transitions in complex environments (2) the model disagreement of the forward and backward models in CABI under different horizons, aiming at checking whether the model disagrees with each other more with the increment of the horizon. To begin with, we define the one-step model prediction error to check whether CABI admits more accurate transitions.

\begin{definition}[Model Prediction Error]
Given the static offline dataset $\mathcal{D}_{\mathrm{env}}$, we define one-step model prediction error for forward model $\epsilon_{\mathrm{fwd}}$ and reverse model $\epsilon_{\mathrm{bwd}}$ as:
\begin{equation*}
\label{eq:modelerror}
\begin{aligned}
\epsilon_{\mathrm{fwd}} = \mathbb{E}_{\substack{(s,a,r,s^\prime)\sim\mathcal{D}_{\mathrm{env}}\\ \hat{s},\hat{r}\sim\hat{p}_\psi(\cdot|s,a)}} \left[ \| s^\prime - \hat{s} \|_2^2 + (r - \hat{r})^2 \right], \\
\epsilon_{\mathrm{bwd}} = \mathbb{E}_{\substack{(s,a,r,s^\prime)\sim\mathcal{D}_{\mathrm{env}}\\ \tilde{s},\tilde{r}\sim\hat{p}_\phi(\cdot|s^\prime,a)}} \left[ \| s - \tilde{s} \|_2^2 + (r - \tilde{r})^2 \right].
\end{aligned}
\end{equation*}
\end{definition}
$\epsilon_{\mathrm{fwd}}$ and $\epsilon_{\mathrm{bwd}}$ generally capture the accuracy of the trained dynamics models, i.e., smaller $\epsilon_{\mathrm{fwd}}$ and $\epsilon_{\mathrm{bwd}}$ indicate better forward and backward dynamics model fitting. 
Intuitively, the one-step model prediction error of admitted samples in CABI should be smaller than that of the mere forward dynamics model or reverse dynamics model, as only transitions that the forward model and backward model are all confident about are admitted. 
We verify this by comparing the one-step model error in the forward model, backward model, and CABI, where we keep the top 20\% imagined samples for CABI. The results are presented in Table \ref{tab:onstespmodelerror}, where we observe CABI leads to significant error drop for both forward and reverse models on all of the tasks. For example, the forward error in \emph{door-cloned} drops from 24.7 to \textbf{0.05} and the backward error drops from 27.7 to \textbf{0.01}, which reveals that CABI can select reliable and conservative imaginations that well fit the dataset for training.

\begin{table}[t]
  \caption{Comparison of one-step model prediction error of the forward model, reverse model, and bidirectional model with the double check mechanism on the Adroit tasks.}
  \label{tab:onstespmodelerror}
  \vspace{0.2cm}
  \renewcommand\arraystretch{1.05}
  \centering
  \small
  \begin{tabular}{@{}lllll@{}}
    \toprule
    \multicolumn{1}{c}{\multirow{2}{*}{Task Name}} & \multicolumn{2}{c}{Unidirectional} & \multicolumn{2}{c}{Bidirectional (CABI)} \\ \cline{2-5}
\multicolumn{1}{c}{}& $\epsilon_{\mathrm{fwd}}$   & $\epsilon_{\mathrm{bwd}}$  & $\epsilon_{\mathrm{fwd}}$   & $\epsilon_{\mathrm{bwd}}$   \\
    \midrule
    pen-cloned & 837.5 & 777.4 & \textbf{751.5} & \textbf{603.0} \\
    pen-human & 195.0 & 177.8 & \textbf{107.5} & \textbf{97.8} \\
    pen-expert & 169.01 & 179.8 & \textbf{143.58} & \textbf{149.8} \\
    door-cloned & 24.7 & 27.7 & \textbf{0.05} & \textbf{0.01} \\
    door-human & 18.2 & 20.2 & \textbf{4.4} & \textbf{6.0} \\
    door-expert & 4.3 & 10.5 & \textbf{1.8} & \textbf{6.3} \\
    relocate-cloned & 351.9 & 1271.4 & \textbf{0.0} & \textbf{0.9} \\
    relocate-human & 229.5 & 267.4 & \textbf{178.6} & \textbf{205.1} \\
    relocate-expert & 201.5 & 48.3 & \textbf{167.5} & \textbf{37.9} \\
    hammer-cloned & 1330.8 & 1984.3 & \textbf{72.3} & \textbf{1602.2} \\
    hammer-human & 577.9 & 596.4 & \textbf{480.9} & \textbf{477.8} \\
    hammer-expert & 601.1 & 561.4 & \textbf{557.2} & \textbf{503.4} \\
    \bottomrule
  \end{tabular}
 \vspace{-0.2cm}
\end{table} 

We then define the model disagreement of forward model and backward model in the following.
\begin{definition}[Bidirectional Model Disagreement]
\label{def:disagreement}
For a sampled current state $s$ and reward $r$ from a given static offline dataset $\mathcal{D}$, a series of forward states $\hat{s}_i$ and reward signals $\hat{r}_i$ can be generated by utilizing the forward model, $i=1,\ldots,H$. Denote the imagined backward state and reward based on $\hat{s}_i$ as $\tilde{s}_{i-1}$ and $\tilde{r}_{i-1}$, respectively, $i=1,\ldots,H$. Then the forward model disagreement is defined as:
\begin{equation}
    \label{eq:forwardmodeldisagreement}
    \epsilon_i^{\mathrm{fwd}} = \begin{cases}
    \mathbb{E}\left[ \| \tilde{s}_0 - s \|_2^2 + (\tilde{r}_0 - r)^2 \right], \quad \mathrm{if} \ i=1, \\
    \mathbb{E}\left[ \| \tilde{s}_{i-1} - \hat{s}_{i-1} \|_2^2 + (\tilde{r}_{i-1} - \hat{r}_{i-1})^2 \right], \quad \mathrm{if} \ i \ge 2.
    \end{cases}
\end{equation}
Similarly, for a sampled next state $s^\prime$ and reward $r$ from the offline dataset, an imaginary trajectory $\hat{\tau}_{\mathrm{bwd}}$ containing the backward states $\tilde{s}_{-i}$ and rewards $\tilde{r}_{-i}$, $i=1,\ldots, H$, can be generated with the aid of the backward dynamics model. For each imagined state $\tilde{s}_{-i}$ in $\hat{\tau}_{\mathrm{bwd}}$, its previous state $\hat{s}_{-i+1}$ and reward $\hat{r}_{-i+1}$ are generated by the forward model, $i=1,\ldots, H$. Then the backward model disagreement is defined as:
\begin{equation}
    \label{eq:backwardmodeldisagreement}
    \epsilon_i^{\mathrm{bwd}} = \begin{cases}
    \mathbb{E}\left[ \| \hat{s}_0 - s^\prime \|_2^2 + (\hat{r}_0 - r)^2 \right], \quad \mathrm{if} \ i=1, \\
    \mathbb{E}\left[ \| \tilde{s}_{-i+1} - \hat{s}_{-i+1} \|_2^2 + (\tilde{r}_{-i+1} - \hat{r}_{-i+1})^2 \right], \quad \mathrm{if} \ i \ge 2.
    \end{cases}
\end{equation}
\end{definition}

\noindent\textbf{Remark:} The above definition generally capture the disagreement between the forward model and the backward model. Note that the model disagreement is different from the model prediction error defined above even if the rollout length is set as 1. The model prediction error measure how well the forward or backward model fits the transition data, while the model disagreement measures how the forward model and backward model disagree on the transition. We take the forward setting as an example. The forward model prediction error is the deviation between the forward imagined state and reward against the real next state and reward signal, while the forward model disagreement is the deviation between the real \emph{current state} and scalar reward with the backward imagined \emph{current state} and reward based on the forward imagination. 

Table \ref{tab:modeldisagreementcabi} details model disagreement comparison of CABI against CABI without double check mechanism, which turns into BOMI, i.e., bidirectional modeling without double check, under different horizons. We perform experiments on 12 Adroit tasks and the sampled mini-batch size is set to be $5\times 10^4$. As demonstrated in the table, the model disagreement of CABI is significantly smaller than that of BOMI under different rollout steps. It is worth noting that the model disagreement for both CABI and BOMI is irrelevant to the rollout length. The model disagreement generally is small when performing one-step model rollout, and increases if longer horizon imaginations are generated (some datasets like \emph{door-human} are exceptions). We observe that the model disagreement in CABI is much more controllable than BOMI, e.g., on some expert datasets.

\begin{table}[!htb]
  \caption{The model disagreement comparison of \textbf{CABI} and \textbf{BOMI} under different rollout length. The superscript $\mathrm{fwd}$ denotes forward, and $\mathrm{bwd}$ denotes backward. The subscript denotes the imagined horizon, e.g., $\epsilon_1^{\mathrm{fwd}}$ represents the forward model disagreement under horizon 1. The best results are in \textbf{bold} (smaller is better).}
  \label{tab:modeldisagreementcabi}
    \vspace{0.2cm}
  \centering
  \setlength{\tabcolsep}{2pt}
  \footnotesize
  \begin{tabular}{lccccccccccccc}
    \toprule
    \multicolumn{1}{c}{\multirow{2}{*}{Task Name}} & \multicolumn{2}{c}{$\epsilon_1^{\mathrm{fwd}}$} & \multicolumn{2}{c}{$\epsilon_2^{\mathrm{fwd}}$} & \multicolumn{2}{c}{$\epsilon_3^{\mathrm{fwd}}$} & \multicolumn{2}{c}{$\epsilon_1^{\mathrm{bwd}}$} & \multicolumn{2}{c}{$\epsilon_2^{\mathrm{bwd}}$} & \multicolumn{2}{c}{$\epsilon_3^{\mathrm{bwd}}$} \\ \cmidrule{2-13}
\multicolumn{1}{c}{}& CABI & BOMI  & CABI   & BOMI & CABI & BOMI & CABI & BOMI & CABI & BOMI & CABI & BOMI   \\
    \midrule
    pen-human & \textbf{0.23} & 1829.21 & \textbf{20.63} & 1608.66 & \textbf{21.24} & 1608.94 & \textbf{0.23} & 1857.14 & \textbf{19.93} & 1693.70 & \textbf{19.25} & 1686.21 \\
    door-human & \textbf{0.53} & 36.03 & \textbf{0.21} & 28.06 & \textbf{0.21} & 29.04 & \textbf{0.53} & 36.19 & \textbf{0.40} & 29.38 & \textbf{0.38} & 30.07  \\
    relocate-human & \textbf{0.00} & 268.37 & \textbf{5.32} & 237.03 & \textbf{5.45} & 236.64 & \textbf{0.00} & 265.34 & \textbf{4.68} & 193.56 & \textbf{4.72} & 195.08 \\
    hammer-human & \textbf{0.67} & 401.58 & \textbf{5.56} & 261.43 & \textbf{5.53} & 259.86 & \textbf{0.69} & 402.67 & \textbf{5.63} & 260.93 & \textbf{5.43} & 258.86 \\
    pen-cloned & \textbf{159.58} & 12048.06 & \textbf{373.10} & 22506.36 & \textbf{359.56} & 22440.26 & \textbf{136.89} & 11624.98 & \textbf{359.91} & 22350.11 & \textbf{331.18} & 21939.83 \\
    door-cloned & \textbf{0.00} & 46.35 & \textbf{0.00} & 63.75 & \textbf{0.00} & 62.89 & \textbf{0.00} & 30.68 & \textbf{0.00} & 61.03 & \textbf{0.00} & 62.96 \\
    relocate-cloned & \textbf{0.00} & 745.77 & \textbf{0.00} & 938.51 & \textbf{0.00} & 929.95 & \textbf{0.00} & 269.66 & \textbf{0.00} & 909.64 & \textbf{0.00} & 889.79 \\
    hammer-cloned & \textbf{0.0} & 3791.10 & \textbf{0.01} & 4509.45 & \textbf{0.01} & 4395.46 & \textbf{0.0} & 1021.85 & \textbf{0.02} & 4443.90 & \textbf{0.03} & 4447.14 \\
    pen-expert & \textbf{0.42} & 1965.76 & \textbf{21.92} & 9642.54 & \textbf{26.05} & 9691.13 & \textbf{0.41} & 1953.54 & \textbf{56.12} & 9569.12 & \textbf{52.69} & 9399.45 \\
    door-expert & \textbf{0.95} & 257.49 & \textbf{0.94} & 261.84 & \textbf{1.04} & 263.52 & \textbf{1.01} & 258.51 & \textbf{0.79} & 276.60 & \textbf{0.69} & 276.12 \\
    relocate-expert & \textbf{0.05} & 649.59 & \textbf{1.47} & 770.99 & \textbf{2.18} & 784.79 & \textbf{0.06} & 652.63 & \textbf{0.06} & 631.83 & \textbf{0.05} & 631.74 \\
    hammer-expert & \textbf{1.03} & 6135.13  & \textbf{58.07} & 82422.41 & \textbf{60.78} & 82633.47 & \textbf{1.02} & 6198.06 & \textbf{11.24} & 92882.87 & \textbf{10.70} & 92102.51 \\
    \bottomrule
  \end{tabular}
\end{table}


The double check mechanism we introduced in the main text selects trustworthy synthetic samples based on the deviation between states, i.e., transition samples with small state deviation will be kept in the model buffer. While we can also trust the transition samples via the model disagreement, i.e., keep transition samples with small model disagreement. We experimentally find that evaluating the deviation between states brings almost the same performance as evaluating the model disagreement under the identical hyperparameter setup. We choose to select transitions according to the deviation between states alone as shown in Algorithm \ref{alg:cabidatageneration} for both space and time saving during data generation process of CABI.

\section{Omitted Background for VAE}
\label{sec:vaebackground}
In this section, we provide a brief introduction to the variational autoencoder (VAE) \cite{Kingma2014AutoEncodingVB}. Given a dataset $X = \left\{ x^{(i)} \right\}_{i=1}^N$, the VAE is trained to generate samples that come from the same distribution as the data points. That is to say, the goal of a VAE is to maximize $p_\theta(X) = \prod_{i=1}^N p_\theta\left(x^{(i)}\right)$, where $\theta$ is the parameter of the approximate maximum-likelihood (ML) or maximum a posterior (MAP) estimation. To reach this goal, a latent variable $z$ sampled from its posterior distribution $p(z)$ is introduced, and we model a decoder $p_\theta(X|z)$ parameterized by $\theta$. However, directly optimizing the marginal likelihood $p_\theta(X) = \int p(z)p_\theta(X|z)dz$ is intractable. Instead, VAE approximates the true posterior $p_\theta(z|X)$ via training an encoder $q_\phi(z|X)$, and we resort to optimizing the evidence lower bound (ELBO) on the log-likelihood of the data as shown in (\ref{eq:vaelowerbound}).
\begin{equation}
    \label{eq:vaelowerbound}
    \max_{\theta,\phi}\log p_\theta(X) \ge \max_{\theta,\phi}\mathbb{E}_{q_\phi(z|X)}[\log p_\theta(X|z)] - D_{\mathrm{KL}}(q_\phi(z|X)\| p_\theta(z)).
\end{equation}
The first term in the right-hand-side of (\ref{eq:vaelowerbound}) denotes the reconstruction loss, where $z$ is sampled from $q_\phi(z|X)$. The second term represents the KL-divergence between the learned encoder of $z$ and its true prior. The encoder $q_\phi(z|X)$ is usually set to be a multivariate Gaussian distribution with mean ${\bf \mu}_\phi$ and variance ${\bf\Sigma}_\phi$. The prior of the latent variable $z$ is set to be a standard multivariate Gaussian distribution. Optimizing the lower bound in (\ref{eq:vaelowerbound}) enables the trained model to generate samples similar to the data distribution. After the VAE is well trained, we sample $z$ from the encoder $q_\phi(z|X)$ and pass it through the decoder $p_\theta(X|z)$ to obtain samples.

In this work, we use the conditional variational autoencoder (CVAE) \cite{Fujimoto2019OffPolicyDR} to model the behavior policy in the dataset. CVAE is a variant of the vanilla VAE, which aims to model $p_\theta(X|Y)$. Similar to the original ELBO of VAE, CVAE optimizes the conditional lower bound as shown in (\ref{eq:cvaelowerbound}).
\begin{equation}
    \label{eq:cvaelowerbound}
    \max_{\theta,\phi}\log p_\theta(X|Y) \ge \max_{\theta,\phi}\mathbb{E}_{q_\phi(z|X,Y)}[\log p_\theta(X|z,Y)] - D_{\mathrm{KL}}(q_\phi(z|X,Y)\| p_\theta(z|Y)).
\end{equation}

\section{Compute Infrastructure}
\label{sec:computinginfrastructure}
In Table \ref{tab:computing}, we list the compute infrastructure that we use to run all of the baseline algorithms and experiments.

\begin{table}[htb]
\caption{Compute infrastructure.}
\label{tab:computing}
\centering
\begin{tabular}{c|c|c}
\toprule
\textbf{CPU}  & \textbf{GPU} & \textbf{Memory} \\
\midrule
AMD EPYC 7452  & RTX3090$\times$8 & 288GB \\
\bottomrule
\end{tabular}
\end{table}

\section{Experimental Results of CABI+IQL}
\label{sec:iql}
In this section, we additionally incorporate CABI with a recently proposed offline RL method, IQL \cite{kostrikov2022offline}. IQL learns without querying OOD samples. Such a learning paradigm ensures that the whole learning process is conducted under the support of the dataset, and a safe policy can be learned. However, as we explained in the main text, the datasets often cannot contain all possible transitions. Hence, the generalization capability of IQL is actually limited. With the aid of CABI, such concern can be mitigated to some extent. We conduct experiments on 12 Adroit datasets and 15 MuJoCo datasets over 5 different random seeds. For IQL, we use its official codebase (\href{https://github.com/ikostrikov/implicit_q_learning}{https://github.com/ikostrikov/implicit\_q\_learning}) to run on all 27 datasets over 5 random seeds with the hyperparameters suggested by the authors. We incorporate CABI with IQL and run CABI+IQL over 5 different random seeds. The forward and backward horizons for CABI+IQL are identical to CABI+BCQ on Adroit tasks and CABI+TD3\_BC on MuJoCo datasets. We summarize the results in Table \ref{tab:iqladroit} and Table \ref{tab:iqlmujoco}. 

As shown, CABI boosts the performance of IQL on all 27 datasets of Adroit and MuJoCo. CABI+IQL outperforms baseline methods on 10 out of 12 datasets. While on MuJoCo datasets, CABI+IQL only surpasses baseline methods on 5 out of 15 datasets, due to the fact that the base method IQL itself has poor performance on "-v0" datasets. Nevertheless, CABI+IQL has a total score of 604.1 on Adroit, surpassing the total score 562.5 of the vanilla IQL. CABI+IQL achieves a total score of 909.3 on MuJoCo datasets, while vanilla IQL only has a total score of 860.7. We want to emphasize here that we do not aim to beat the most recent strong baseline methods in this paper, the key point we want to carry here is the conservative data augmentation with CABI is effective and beneficial for the performance improvement over the base offline RL algorithms. The empirical experiments work as the evidence to validate our claim.

\begin{table*}[!t]
  \caption{Normalized average score comparison of CABI+IQL against different baselines on the Adroit "-v0" tasks, where score 0 represents the performance of a random policy and 100 corresponds to an expert policy performance. The highest mean scores are in \textbf{bold}.}
  \vspace{0.2cm}
  \renewcommand\tabcolsep{4.5pt}
  \renewcommand\arraystretch{1.05}
  \label{tab:iqladroit}
  \centering
  \footnotesize
  \begin{tabular}{@{}llllllllll@{}}
    \toprule
    Task Name  & CABI+IQL & IQL & UWAC & BEAR & BC & AWR & CQL & MOPO & COMBO \\
    \midrule
    pen-cloned  & 42.2$\pm$6.1 & 35.2$\pm$7.3 & 33.1 & 26.5 & \textbf{56.9} & 28.0 & 39.2 & -2.1 & -2.4\\
    pen-human & \textbf{72.0}$\pm${9.1} & 68.7$\pm$8.6 & 21.7 & -1.0 & 34.4 & 12.3 & 37.5 & 9.7  & 27.7\\
    pen-expert & \textbf{129.1}$\pm${0.6} & 118.4$\pm$6.9 & 111.9 & 105.9 & 85.1 & 111.0 & 107.0 & -0.6 & 11.5  \\
    door-cloned & \textbf{0.8}$\pm$0.4 & 0.7$\pm$0.5 & 0.0 & -0.1 & -0.1 & 0.0 & 0.4 & -0.1  & 0.0   \\
    door-human & \textbf{11.5}$\pm$3.6 & 3.3$\pm$1.3 & 2.1 & -0.3 & 0.5 & 0.4 & 9.9 & -0.2  & -0.3 \\
    door-expert & \textbf{105.7}$\pm${0.2} & 105.2$\pm$0.3 & 104.1 & 103.4 & 34.9 & 102.9 & 101.5 & -0.2  & 4.9  \\
    relocate-cloned & \textbf{-0.1}$\pm$0.0 & -0.2$\pm$0.0 & -0.3 & -0.3 & \textbf{-0.1} & -0.2 & \textbf{-0.1} & -0.3  & \textbf{-0.1}   \\
    relocate-human & 0.4$\pm$0.3 & 0.0$\pm$0.0 & \textbf{0.5} & -0.3 & 0.0 & 0.0 & 0.2 & -0.3  & -0.3  \\
    relocate-expert & \textbf{107.4}$\pm${0.2} & 105.6$\pm$0.5 & 105.6 & 98.6 & 101.3 & 91.5 & 95.0 & -0.2  & 17.2  \\
    hammer-cloned & \textbf{2.4} $\pm${0.2} & 1.6$\pm$1.0 & 0.4 & 0.3 & 0.8 & 0.4 & 2.1 & 0.2 & 0.4  \\
    hammer-human & \textbf{4.8}$\pm$1.8 & 2.3$\pm$0.6 & 1.1 & 0.3 & 1.5 & 1.2 & 4.4 & 0.2 & 0.2   \\
    hammer-expert & \textbf{127.9}$\pm${1.2} & 121.7$\pm$1.3 & 110.6 & 127.3 & 125.6 & 39.0 & 86.7 & 0.3  & 0.3  \\
    \midrule
    Total Score & \textbf{604.1} & 562.5 & 490.8 & 460.3 & 440.8 & 386.5 & 483.8 & 6.4 & 59.1 \\
    \bottomrule
  \end{tabular}
\end{table*}

\begin{table*}[t]
  \caption{Normalized average score comparison of CABI+IQL vs. baseline methods on the D4RL MuJoCo "-v0" dataset, where score 0 corresponds to a random policy performance and 100 corresponds to an expert policy performance. The highest mean scores are in \textbf{bold}.}
  \label{tab:iqlmujoco}
\vspace{0.2cm}
\renewcommand\tabcolsep{3pt}
  \renewcommand\arraystretch{1.05}
  \centering
  \footnotesize
  \begin{tabular}{@{}llllllllll@{}}
    \toprule
    Task Name & CABI+IQL & IQL & UWAC & MOPO & BCQ & BC & CQL & FisherBRC \\
    \midrule
    halfcheetah-random & 18.4$\pm$0.9 & 16.2$\pm$0.2 & 2.3 & \textbf{35.4} & 2.2 & 2.0 & 21.7 & 32.2 \\
    hopper-random & 11.4$\pm$0.1 & 9.3$\pm$1.8 & 9.8 & \textbf{11.7} & 10.6 & 9.5 & 10.7 & 11.4 \\
    walker2d-random & 8.0$\pm$0.5 & 6.2$\pm$2.2 & 3.8 & \textbf{13.6} & 4.9 & 1.2 & 2.7 & 0.6 \\
    halfcheetah-medium-replay & 42.2$\pm$0.2 & 40.5$\pm$0.4 & 38.9 & \textbf{53.1} & 38.2 & 34.7 & 41.9 & 43.3 \\
    hopper-medium-replay & 36.8$\pm$0.4 & 33.4$\pm$1.1 & 18.0 & \textbf{67.5} & 33.1 & 19.7 & 28.6 & 35.6 \\
    walker2d-medium-replay & 17.2$\pm$0.8 & 15.8$\pm$1.7 & 8.4 & 39.0 & 15.0 & 8.3 & 15.8 & \textbf{42.6} \\
    halfcheetah-medium & 41.6$\pm${0.1} & 41.2$\pm$0.1 & 37.4 & \textbf{42.3} & 40.7 & 36.6 & 37.2 & 41.3 \\
    hopper-medium & 40.0$\pm$12.9 & 30.7$\pm$0.0 & 30.3 & 28.0 & 54.5 & 30.0 & 44.2 & \textbf{99.4} \\
    walker2d-medium & 55.1$\pm$2.3 & 50.8$\pm$7.7 & 17.4 & 17.8 & 53.1 & 11.4 & 57.5 & \textbf{79.5} \\
    halfcheetah-medium-expert & \textbf{96.7}$\pm${1.3} & 89.0$\pm$0.7 & 40.6 & 63.3 & 64.7 & 67.6 & 27.1 & 96.1 \\
    hopper-medium-expert & \textbf{112.8}$\pm${0.2} & 111.5$\pm$1.0 & 95.4 & 23.7 & 110.9 & 89.6 & 111.4 & 90.6 \\
    walker2d-medium-expert & \textbf{104.8}$\pm${1.0} & 99.7$\pm$2.9 & 14.8 & 44.6 & 57.5 & 12.0 & 68.1 & 103.6 \\
    halfcheetah-expert & 104.7$\pm$0.9 & 100.8$\pm$3.7 & 104.0 & - & 89.9 & 105.2 & 82.4 & \textbf{106.8} \\
    hopper-expert & \textbf{112.8}$\pm${0.2} & 112.0$\pm$0.0 & 109.1 & - & 107.0 & 111.5 & 111.2 & 112.3 \\
    walker2d-expert & \textbf{106.8}$\pm${3.7} & 103.6$\pm$2.0 & 88.4 & - & 102.3 & 56.0 & 103.8 & 79.9 \\
    \midrule
    Total Score & 909.3 & 860.7 & 618.6 & - & 784.6 & 595.3 & 764.3 & \textbf{974.6}\\
    \bottomrule
  \end{tabular}
\end{table*}

\section{Omitted Full Comparison of CABI against Baselines}
\label{sec:fullcomptable}
In this section, we provide the full comparison of CABI against baseline methods as we omit standard deviation for baselines in the main text due to space limitation. We show in Table \ref{tab:adroitfullscorecompare} the full performance comparison of CABI+BCQ against BCQ \cite{Fujimoto2019OffPolicyDR}, UWAC \cite{Wu2021UncertaintyWA}, CQL \cite{Kumar2020ConservativeQF}, MOPO \cite{Yu2020MOPOMO}, COMBO \cite{Yu2021COMBOCO}, etc. We additionally compare against SAC. As MOPO and COMBO do not report the performance on the Adroit dataset in their original paper, we run COMBO on the Adroit tasks over 5 different random seeds with our reproduced code, and run MOPO and UWAC with their official codebases on the Adroit tasks over 5 different random seeds, respectively. The results of the rest of the baselines are taken directly from \cite{Fu2020D4RLDF}. 

Table \ref{tab:mujocofullscorecompare} gives the full comparison of CABI+TD3\_BC against TD3\_BC \cite{Fujimoto2021AMA}, BCQ \cite{Fujimoto2019OffPolicyDR}, UWAC \cite{Wu2021UncertaintyWA}, FisherBRC \cite{Kostrikov2021OfflineRL}, CQL \cite{Kumar2020ConservativeQF}, MOPO \cite{Yu2020MOPOMO}, etc. We additionally compare CABI+BCQ against BEAR here. The results of BC, TD3\_BC, CQL, and FisherBRC are taken directly from \cite{Fujimoto2021AMA}, and the results of UWAC are acquired by running the official codebase over 5 different random seeds. The results of BEAR are taken directly from \cite{Wu2021UncertaintyWA}. We do not report standard deviation for BEAR and BCQ.

\begin{sidewaystable}[!t]
  \caption{Normalized average score comparison of CABI+BCQ against different baselines on the Adroit tasks, where score 0 represents the performance of a random policy and 100 corresponds to an expert policy performance. The highest mean scores are in \textbf{bold}.}
  \vspace{0.2cm}
  \renewcommand\arraystretch{1.05}
  \label{tab:adroitfullscorecompare}
  \centering
  \begin{tabular}{@{}llllllllllll@{}}
    \toprule
    Task Name  & CABI+BCQ & BCQ & UWAC & BEAR & BC & AWR & CQL & SAC-off & MOPO & COMBO \\
    \midrule
    pen-cloned  & 54.7$\pm$2.0 & 44.0 & 33.1$\pm$10.1 & 26.5 & \textbf{56.9} & 28.0 & 39.2 & 23.5 & -2.1$\pm$4.0 & -2.4$\pm$1.1 \\
    pen-human & \textbf{75.1}$\pm${1.5} & 68.9& 21.7$\pm$2.2 & -1.0 & 34.4 & 12.3 & 37.5 & 6.3 & 9.7$\pm$4.1  & 27.7$\pm$10.7 \\
    pen-expert & \textbf{127.6}$\pm${2.0} & 114.9 & 111.9$\pm$1.2 & 105.9 & 85.1 & 111.0 & 107.0 & 6.1 & -0.6$\pm$1.9 & 11.5$\pm$2.3  \\
    door-cloned & \textbf{0.5}$\pm$0.2 & 0.0 & 0.0$\pm$0.0 & -0.1 & -0.1 & 0.0 & 0.4 & 0.0 & -0.1$\pm$0.1  & 0.0$\pm$0.0  \\
    door-human & 1.7$\pm$0.1 & -0.0 & 2.1$\pm$1.6 & -0.3 & 0.5 & 0.4 & \textbf{9.9} & 3.9 & -0.2$\pm$0.1  & -0.3$\pm$0.2 \\
    door-expert & \textbf{105.3}$\pm${0.5} & 99.0 & 104.1$\pm$1.8 & 103.4 & 34.9 & 102.9 & 101.5 & 7.5 & -0.2$\pm$0.1  & 4.9$\pm$1.2 \\
    relocate-cloned & -0.2$\pm$0.0 & -0.3 & -0.3$\pm$0.0 & -0.3 & \textbf{-0.1} & -0.2 & \textbf{-0.1} & -0.2 & -0.3$\pm$0.1  & \textbf{-0.1}$\pm$0.1  \\
    relocate-human & 0.1$\pm$0.1 & \textbf{0.5} & \textbf{0.5}$\pm$0.6 & -0.3 & 0.0 & 0.0 & 0.2 & 0.0 & -0.3$\pm$0.1  & -0.3$\pm$0.1 \\
    relocate-expert & \textbf{105.9}$\pm${1.0} & 41.6 & 105.6$\pm$1.6 & 98.6 & 101.3 & 91.5 & 95.0 & -0.3 & -0.2$\pm$0.0  & 17.2$\pm$3.1 \\
    hammer-cloned & \textbf{4.3} $\pm${1.6} & 0.4 & 0.4$\pm$0.0 & 0.3 & 0.8 & 0.4 & {2.1} & 0.2 & 0.2$\pm$0.0 & 0.4$\pm$0.2 \\
    hammer-human & 3.1$\pm$2.2 & 0.5 & 1.1$\pm$0.6 & 0.3 & 1.5 & 1.2 & \textbf{4.4} & 0.5 & 0.2$\pm$0.0 & 0.2$\pm$0.0  \\
    hammer-expert & \textbf{128.9}$\pm${0.9} & 107.2 & 110.6$\pm$20.7 & 127.3 & 125.6 & 39.0 & 86.7 & 25.2 & 0.3$\pm$0.0  & 0.3$\pm$0.1 \\
    \midrule
    Total Score & \textbf{607.0} & 476.7 & 490.8 & 460.3 & 440.8 & 386.5 & 483.8 & 72.7 & 6.4 & 59.1 \\
    \bottomrule
  \end{tabular}
\end{sidewaystable}

\begin{sidewaystable}[t]
  \caption{Normalized average score of CABI+TD3\_BC vs. prior state-of-the-art methods on the D4RL MuJoCo dataset, where score 0 corresponds to a random policy performance and 100 corresponds to an expert policy performance.}
  \label{tab:mujocofullscorecompare}
\vspace{0.2cm}
\renewcommand\tabcolsep{5pt}
  \renewcommand\arraystretch{1.05}
  \centering
  \begin{tabular}{@{}lllllllllll@{}}
    \toprule
    Task Name & CABI+TD3\_BC & TD3\_BC & UWAC & MOPO & BEAR & BCQ & BC & CQL & FisherBRC \\
    \midrule
    halfcheetah-random & 15.1$\pm$0.4 & 10.2$\pm$1.3 & 2.3$\pm$0.0 & \textbf{35.4}$\pm$2.5 & 25.1 & 2.2 & 2.0$\pm$0.1 & 21.7$\pm$0.9 & 32.2$\pm$2.2 \\
    hopper-random & \textbf{11.9}$\pm$0.1 & 11.0$\pm$0.1 & 9.8$\pm$0.1 & 11.7$\pm$0.4 & 11.4 & 10.6 & 9.5$\pm$0.1 & 10.7$\pm$0.1 & 11.4$\pm$0.2 \\
    walker2d-random & 6.4$\pm$1.5 & 1.4$\pm$0.6 & 3.8$\pm$1.3 & \textbf{13.6}$\pm$2.6 & 7.3 & 4.9 & 1.2$\pm$0.2 & 2.7$\pm$1.2 & 0.6$\pm$0.6 \\
    halfcheetah-med-replay & 44.4$\pm$0.2 & 43.3$\pm$0.5 & 38.9$\pm$0.3 & \textbf{53.1}$\pm$2.0 & 38.6 & 38.2 & 34.7$\pm$1.8 & 41.9$\pm$1.1 & 43.3$\pm$0.9 \\
    hopper-med-replay & 31.3$\pm$0.7 & 31.4$\pm$3.0 & 18.0$\pm$5.6 & \textbf{67.5}$\pm$24.7 & 33.7 & 33.1 & 19.7$\pm$5.9 & 28.6$\pm$0.9 & 35.6$\pm$2.5 \\
    walker2d-med-replay & 29.4$\pm$1.3 & 25.2$\pm$5.1 & 8.4$\pm$1.2 & 39.0$\pm$9.6 & 19.2 & 15.0 & 8.3$\pm$1.5 & 15.8$\pm$2.6 & \textbf{42.6}$\pm$7.0 \\
    halfcheetah-medium & \textbf{45.1}$\pm${0.1} & 42.8$\pm$0.3 & 37.4$\pm$0.2 & 42.3$\pm$1.6 & 41.7 & 40.7 & 36.6$\pm$0.6 & 37.2$\pm$0.3 & 41.3$\pm$0.5 \\
    hopper-medium & \textbf{100.4}$\pm${0.3} & 99.5$\pm$1.0 & 30.3$\pm$0.3 & 28.0$\pm$12.4 & 52.1 & 54.5 & 30.0$\pm$0.5 & 44.2$\pm$10.8 & 99.4$\pm$0.4 \\
    walker2d-medium & \textbf{82.0}$\pm${0.4} & 79.7$\pm$1.8 & 17.4$\pm$8.5 & 17.8$\pm$19.3 & 59.1 & 53.1 & 11.4$\pm$6.3 & 57.5$\pm$8.3 & 79.5$\pm$1.0 \\
    halfcheetah-med-expert & \textbf{105.0}$\pm${0.2} & 97.9$\pm$4.4 & 40.6$\pm$6.6 & 63.3$\pm$38.0 & 53.4 & 64.7 & 67.6$\pm$13.2 & 27.1$\pm$3.9 & 96.1$\pm$9.5 \\
    hopper-med-expert & \textbf{112.7}$\pm${0.0} & 112.2$\pm$0.2 & 95.4$\pm$23.5 & 23.7$\pm$6.0 & 96.3 & 110.9 & 89.6$\pm$27.6 & 111.4$\pm$1.2 & 90.6$\pm$43.3 \\
    walker2d-med-expert & \textbf{108.4}$\pm${1.3} & 101.1$\pm$9.3 & 14.8$\pm$1.4 & 44.6$\pm$12.9 & 40.1 & 57.5 & 12.0$\pm$5.8 & 68.1$\pm$13.1 & 103.6$\pm$4.6 \\
    halfcheetah-expert & \textbf{107.6}$\pm$0.9 & 105.7$\pm$1.9 & 104.0$\pm$2.0 & - & \textbf{108.2} & - &  105.2$\pm$1.7 & 82.4$\pm$7.4 & 106.8$\pm$3.0 \\
    hopper-expert & \textbf{112.4}$\pm${0.1} & 112.2$\pm$0.2 & 109.1$\pm$3.9 & - & 110.3 & - & 111.5$\pm$1.3 & 111.2$\pm$2.1 & 112.3$\pm$0.2 \\
    walker2d-expert & \textbf{108.6}$\pm${1.5} & 105.7$\pm$2.7 & 88.4$\pm$3.7 & - & 106.1 & - & 56.0$\pm$24.9 & 103.8$\pm$7.6 & 79.9$\pm$32.4 \\
    \midrule
    Total Score & \textbf{1020.7} & 979.3 & 618.6 & - & 802.6 & - & 595.3 & 764.3 & 974.6\\
    \bottomrule
  \end{tabular}
\end{sidewaystable}

\end{document}